\documentclass[journal]{IEEEtai}
\usepackage{amsmath,amsfonts}
\usepackage{bbm}
\usepackage{algorithmic}
\usepackage{array}
\usepackage{textcomp}
\usepackage{stfloats}
\usepackage{url}
\usepackage{verbatim}
\usepackage{cite}
\usepackage{url}
\usepackage{arydshln}
\usepackage{booktabs}
\usepackage[flushleft]{threeparttable}
\usepackage{color}
\usepackage{listings}
\usepackage{amssymb}
\usepackage{graphicx}
\usepackage{subfig}
\usepackage{multicol, blindtext}
\usepackage{multirow}
\usepackage{makecell}
\usepackage{float}
\usepackage{multirow}
\usepackage{longtable,lscape}
\usepackage{cases}
\usepackage{diagbox}
\usepackage[normalem]{ulem}
\usepackage[colorlinks]{hyperref}
\usepackage[absolute]{textpos}

\usepackage{url}
\usepackage{caption}
\usepackage{color}
\usepackage{xcolor}
\usepackage{amsthm}
\usepackage[pagewise]{lineno}

\usepackage{nomencl}
\makenomenclature

\usepackage{etoolbox}
\renewcommand\nomgroup[1]{%
 \item[\bfseries
  \ifstrequal{#1}{A}{States and Actions}{%
  \ifstrequal{#1}{B}{Car-following Models}{%
  \ifstrequal{#1}{C}{Other Notations}{}}}%
]}

\usepackage{array}
\usepackage{multirow}
\usepackage{multicol}
\usepackage{makecell}
\usepackage{comment}
\usepackage{graphicx}
\usepackage{soul}

\usepackage{booktabs,caption}
\usepackage[flushleft]{threeparttable}
\usepackage{arydshln}
\usepackage[ruled,vlined, linesnumbered]{algorithm2e}

\usepackage{enumerate}

\usepackage{makecell}

\theoremstyle {plain}

\newcounter{x}

\theoremstyle{definition}
\newtheorem{defn}{Definition}[section]

\theoremstyle{remark}
\newtheorem*{rem}{Remark}

\newenvironment{eqnsize}
    {\begin{small}}
    {\end{small}}

\usepackage{titlesec}

\usepackage{color}

\newcommand{\tcb}[1]{{\textcolor{blue}{#1}}}

\newcommand{\tcck}[1]{{\textcolor{black}{#1}}}

\begin{document}
%

\begin{textblock}{17}(1.5,0.1)
\noindent \tcb{
Published in \emph{Algorithms}, 2023.
Camera-ready version available at \href{https://www.mdpi.com/1999-4893/16/6/305}{https://www.mdpi.com/1999-4893/16/6/305}. \\
Please cite this paper as: \\
Di, X., Shi, R., Mo, Z. and Fu, Y., 2023. Physics-Informed Deep Learning For Traffic State Estimation: \\
A Survey and the Outlook. Algorithms, 16(6), p.305.
}
\end{textblock}

\title{Physics-Informed Deep Learning For Traffic State Estimation: 
A Survey and the Outlook}

\author{Xuan Di,~\IEEEmembership{Member,~IEEE}, Rongye Shi,~\IEEEmembership{Member,~IEEE}, Zhaobin Mo and Yongjie Fu
\thanks{This work is supported by NSF CPS-2038984. 
\textit{(Corresponding author: Xuan Di.)}}
\thanks{Xuan Di, Rongye Shi, Zhaobin Mo, and Yongjie Fu  are with the Department of Civil Engineering and Engineering Mechanics, Columbia University, New York, NY, 10027 USA (e-mail:  sharon.di@columbia.edu; rongyes@alumni.cmu.edu; zm2302@columbia.edu; yf2578@columbia.edu).} 
\thanks{Xuan Di is also with the Data Science Institute, Columbia University, New York, NY, 10027 USA.}
\thanks{Rongye Shi and Zhaobin Mo contributed equally.}
}

\markboth{}
{Di \MakeLowercase{\textit{et al.}}}
\maketitle



\begin{abstract}
    For its robust predictive power (compared to pure physics-based models) and sample-efficient training (compared to pure deep learning models), physics-informed deep learning (PIDL), a paradigm hybridizing physics-based models and deep neural networks (DNN), has been booming in science and engineering fields. 
    One key challenge of applying PIDL to various domains and problems lies in the design of a computational graph that integrates physics and DNNs. In other words, how physics are encoded into DNNs and how the physics and data components are represented.  
    In this paper, we provide a variety of architecture designs of PIDL computational graphs and how these structures are customized to traffic state estimation (TSE), 
    a central problem in transportation engineering. 
    When observation data, problem type, and goal vary, we demonstrate potential architectures of PIDL computational graphs and compare these variants using the same real-world dataset. 
\end{abstract}
\begin{IEEEImpStatement}
Despite benefits of physics-informed deep learning (PIDL) for robust and efficient pattern learning, 
how to inject physics into neural networks (NN) remains under-explored and varies case by case and across domains.
We will summarize and establish a systematic design pipeline for hybrid computational graphs that facilitates the integration of physics and NNs.
The insights into the modular design of the hybrid PIDL paradigm as well as the established visualization tool will not only be useful to guide transportation researchers to pursue PIDL, but can also facilitate researchers at large to better understand and decompose a PIDL problem when applied to their own application domains.
It will hopefully  open up the conversation across domains about establishing a unified PIDL paradigm. 
\end{IEEEImpStatement}


%
\IEEEpeerreviewmaketitle
\vspace{-1em}
\section{Introduction}
\label{sec:intro}

\begin{figure}[h!]
\centering
  \includegraphics[scale=0.4]{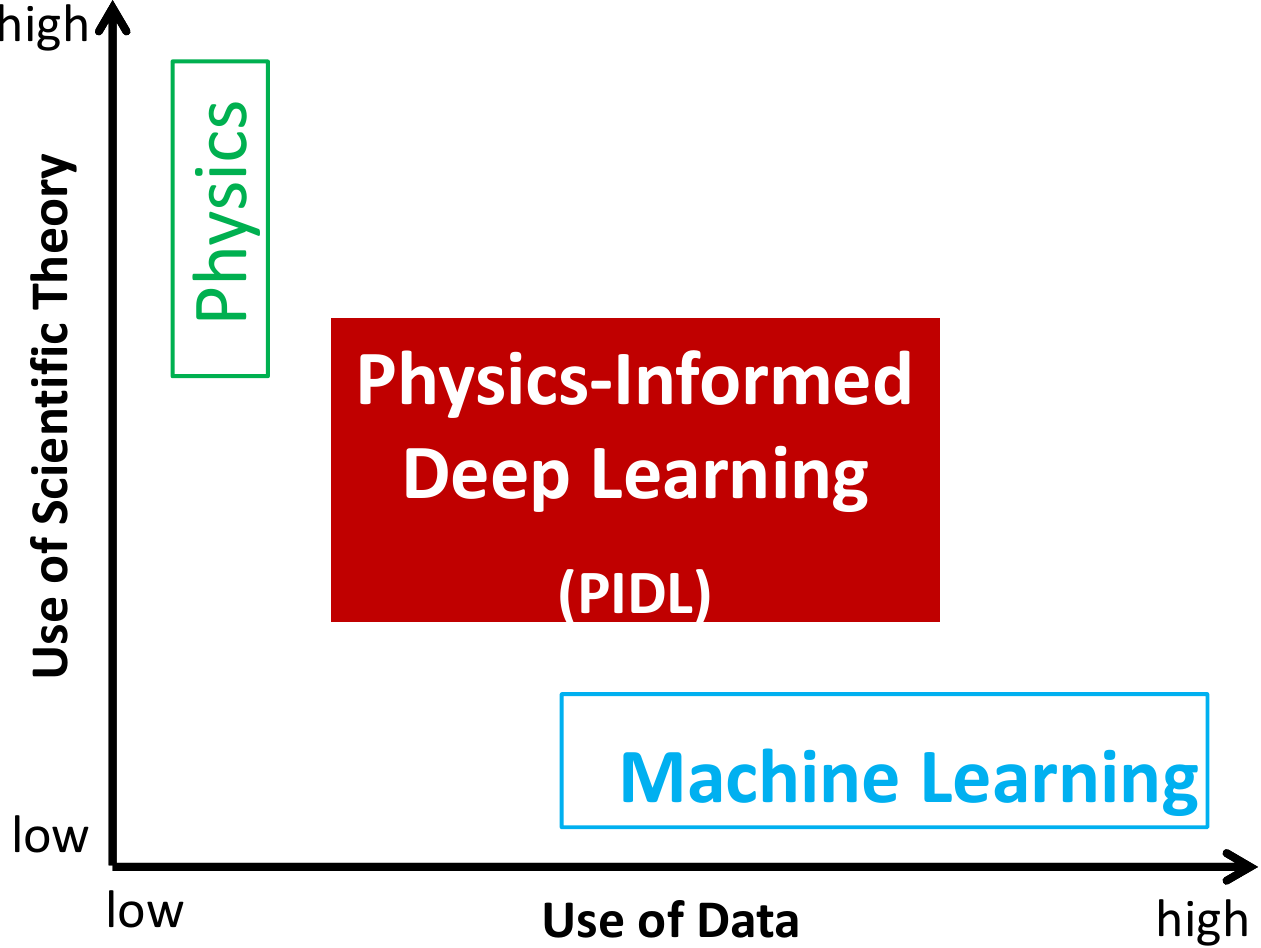}
  \caption{Comparison of the pure physics based, the data-driven, and the hybrid paradigms (adapted from \cite{Karpatne-2017}).}
  \label{fig:pidl_framework}
\end{figure}

\IEEEPARstart{P}{hysics}-informed deep learning (PIDL) \cite{raissi2019physics}, 
also named ``theory-guided data science" \cite{Karpatne-2017}, ``model-informed machine learning" \cite{yu2021model}, or ``physics-informed machine learning" \cite{karniadakis2021physics}, has gained increasing traction in various scientific and engineering fields. 
Its underlying rationale is to leverage the pros of both physics-based and data-driven approaches while compensating the cons of each. 
Physics-based approach refers to scientific hypotheses of what underlying physics govern observations, like the first principle. 
Normally, scientists or engineers first come up with a prior assumption of how a quantity of interest is computed from other physics quantities. 
Then laboratory or field experiments are designed to collect data that are used to calibrate the involved parameters. 
In contrast, data-driven approach does not bear any prior knowledge of how things work and how different quantities are correlated. Instead, they rely on machine learning (ML) techniques such as deep learning (DL) to learn and infer patterns from data. 
The former is data-efficient and interpretable but may not be generalizable to unobserved data, 
while the latter is generalizable at cost of relying on huge amounts of training samples and may be incapable of offering deductive insights.
Thus, the PIDL paradigm opens up a promising research direction that leverages the strengths of both physics-based and data-driven approaches. 
Fig.~\ref{fig:pidl_framework} summarizes the amounts of data (in x-axis) and scientific theory (in y-axis) used for each paradigm. 
Model based approaches heavily rely on the scientific theory discovered from the domain knowledge and little data for system discovery, 
while machine learning approaches mostly rely on data for mechanism discovery, 
In contrast, PIDL employs a small amount of data (i.e., ``small data" \cite{karniadakis2021physics}) for pattern discovery while leveraging a certain amount of scientific knowledge to impose physically consistent constraints. 
The ``sweetspot" of PIDL lies in the small data and partial knowledge regime. 
In other words, PIDL achieves the best performance in accuracy and robustness with small data and partial domain knowledge. 
In this paper, we will validate the advantage of PIDL using transportation applications, and present a series of experiments using the same real-world dataset against conventional physics-based models.

There could be a variety of structure designs for PIDL.
Here we primarily focus on the PIDL framework that is represented by a \emph{hybrid computational graph} (HCG) consisting of two graphs: 
a physics-informed computational graph (PICG)  
and a physics-uninformed neural networks (PUNN).
Despite numerous benefits of PIDL, there are still many open questions that need to be answered,
including 
the construction of the HCG, 
the choice of the physics and the architecture of the PICG, 
the architecture of the PUNN, 
the loss function, and the training algorithms. 
Among them, how to encode physics into (deep) neural networks (DNN) remains under-explored and varies case by case and across domains. 
In this paper, we will primarily establish a systematic design pipeline for hybrid computational graphs that would facilitate the integration of physics and DL. 
To this end, we will review the state-of-the-art of the PIDL architecture in traffic state estimation (TSE) problem. 
This survey paper can be used as a guideline for researchers at large when they consider using PIDL for problems at hand. 

We have seen a growing number of studies that apply PIDL to physical and biological systems \cite{karniadakis2021physics,cuomo2022scientific}. 
However, its feasibility for social systems, in particular, human decision making processes, such as driving behaviors, remains largely unexploited. 
Human's decision-making involves complex cognitive processes compounded by perception errors, noisy observations, and output randomness. 
Moreover, human driving behaviors exhibit highly unstable and nonlinear patterns, leading to stop-and-go traffic waves and traffic congestion \cite{di2021survey,huang2019stable,huang2020stable}. 
Accordingly, neither model-driven nor data-driven approach alone suffices to predict such behaviors with high accuracy and robustness. 
Thereby, we strongly believe that a hybrid method, which leverages the advantage of both model-driven and data-driven approaches, are promising \cite{mo2021physics, mo2022uncertainty}.



There is a vast amount of literature  
for TSE \cite{wang2005real,Seo-17} and 
for PIDL \cite{karniadakis2021physics,cuomo2022scientific}, respectively. 
To distinguish from other survey papers in TSE,
here we primarily focus on data-driven approaches, PIDL in particular. 
Since PIDL for TSE is less studied than physics-based models and the existing literature is focused on single roads, we will primarily  examine TSE along links and leave that on a road network, which includes both link and node models, for future research.
To distinguish from other PIDL surveys, we primarily focus on the modular design of the hybrid PIDL paradigm and show how to customize various designs for accurate and robust identification of traffic dynamic patterns. 
Here the \emph{modular design} refers to the architecture design of each component in the graph and how these components are wired, 
in other words, how physics laws are injected into DNNs. 
The generic architecture of a PIDL consists of two computational graphs: 
one DNN (i.e., the data-driven component) for predicting the unknown solution, 
while the other (i.e., the physics-driven component), in which physics 
are represented, for evaluating whether the prediction aligns with the given physics. 
The physics encoded computational graph can be treated as a regularization term of the other deep neural network to prevent overfitting, i.e., high-variance. 
In summary, the hybrid of both components overcomes high-bias and high-variance induced by each individual one, rendering it possible to leverage the advantage of both the physics-based and data-driven methods in terms of model accuracy and data efficiency.

Application of PIDL to TSE is a relatively new area.
We hope that the insights into the modular design of the hybrid PIDL paradigm as well as the established visualization tool
will not only be useful to guide transportation researchers to pursue PIDL, but also facilitate researchers at large to better understand a PIDL pipeline when applied to their own application domains.

Overall, this paper offers a comprehensive overview of the state-of-the-art in TSE using PIDL, while striving to provide insights into the pipeline of implementing PIDL, from architecture design, training, to testing.
In particular,
\begin{enumerate}
    \item propose a computational graph that visualizes both physics and data components in PIDL;
    \item establish a generic way of designing each module of the PIDL computational graphs for both predication and uncertainty quantification;   
    \item benchmark the performance of various PIDL models using the same real-world dataset and identify the advantage of PIDL in the ``small data" regime. 
\end{enumerate}

The rest of the paper is organized as follows: 
Section \ref{sec:related_works} introduces the preliminaries of TSE and its state-of-the-art. 
Section~\ref{subsec:prob_stat} lays out the framework of PIDL for TSE. 
Two types of problems for TSE, namely, deterministic prediction and uncertainty quantification, are detailed in Sections \ref{sec:pidl_deterministic}-\ref{sec:pidl_uq}, respectively.
Section \ref{sec:conclusion} concludes our work and projects future research directions in this promising arena.

\section{Preliminaries and Related Work}
\label{sec:related_works}

\subsection{PIDL}
\begin{defn}
\textbf{Generic framework for physics-informed deep learning.} 
Define location $x\in [0,L]$ and time $t \in [0,T]$ and $L, T\in\mathbb{R}^+$. 
Then the spatiotemporal (ST) domain of interest is a continuous set of points: ${\cal D}=\{(x,t)|x \in[0,L],t\in[0,T]\}$. 
Denote the state as $\mathbf{s}$ and its observed quantity as $\hat{\mathbf{s}}$.
Denote the (labeled) observation ${\cal O},{\cal B},{\cal I}$ and the (unlabeled) collocation points ${\cal C}$ below:
\vspace{-0.3em}
\begin{equation}
\begin{eqnsize}
\label{equ-3-x1}
\left\{ \hspace{-0.6em} {\begin{array}{*{20}l}
    \  {\cal O}=\{(\mathbf{x}^{(i)}, t^{(i)}; \hat{\mathbf{s}}^{(i)})\}_{i=1}^{N_o}: \textit{within-domain observation}, \vspace{0.3em} \ \  \\
   \ {\cal B}=\{t^{(i_b)}; \hat{\mathbf{s}}^{(i_b)}\}_{i_b=1}^{N_{b}}: \textit{boundary observation}, \vspace{0.3em} \ \ \\
   \ {\cal I}=\{\mathbf{x}^{(i_0)}; \hat{\mathbf{s}}^{(i_0)}\}_{i_0=1}^{N_{0}}: \textit{initial observation}, \vspace{0.3em} \ \ \\
   \ {\cal C}=\{(\mathbf{x}^{(j)}, t^{(j)})\}_{j=1}^{N_c}: \textit{collocation points},
\end{array}} \right.\
\end{eqnsize}
\vspace{-0.3em}
\end{equation}
where, $i$ and $j$ are the indices of observation and collocation points, respectively. 
$i_b,i_0$ are the indices of boundary and initial data, respectively.
The numbers of observed data, boundary and initial conditions, and collocation states are denoted as $N_o, N_{b}, N_{0}, N_c$, respectively.
The subscripts $b,0$ represents boundary and initial condition indices, respectively. 

We design a \emph{hybrid computational graph} (HCG) consisting of two computational graphs: 
(1) a PUNN, denoted as $f_{\theta}(\mathbf{x},t)$, to approximate mapping $\mathbf{s}^{(i)}$, 
and (2) a PICG, denoted as $f_{\lambda}(\mathbf{x},t)$, for computing traffic states of $\mathbf{s}^{(j)}$ from collocation points. 
In summary, a general PIDL model, denoted as $f_{\theta,\lambda}(\mathbf{x},t)$, is to train an optimal parameter set $\theta^*$ for PUNN and an optimal parameter set $\lambda^*$ for the physics. 
The PUNN parameterized by the solution $\theta^{*}$ can then be used to predict a traffic state $\hat{\mathbf{s}}_{new}$ on a new set of observed ST points ${\cal O}_{new}\subseteq {\cal D}$, and $\lambda^{*}$ is the most likely model parameters that describe the intrinsic physics of observed data. 
\end{defn}

\begin{figure}[h!]
\centering
  \includegraphics[scale=0.35]{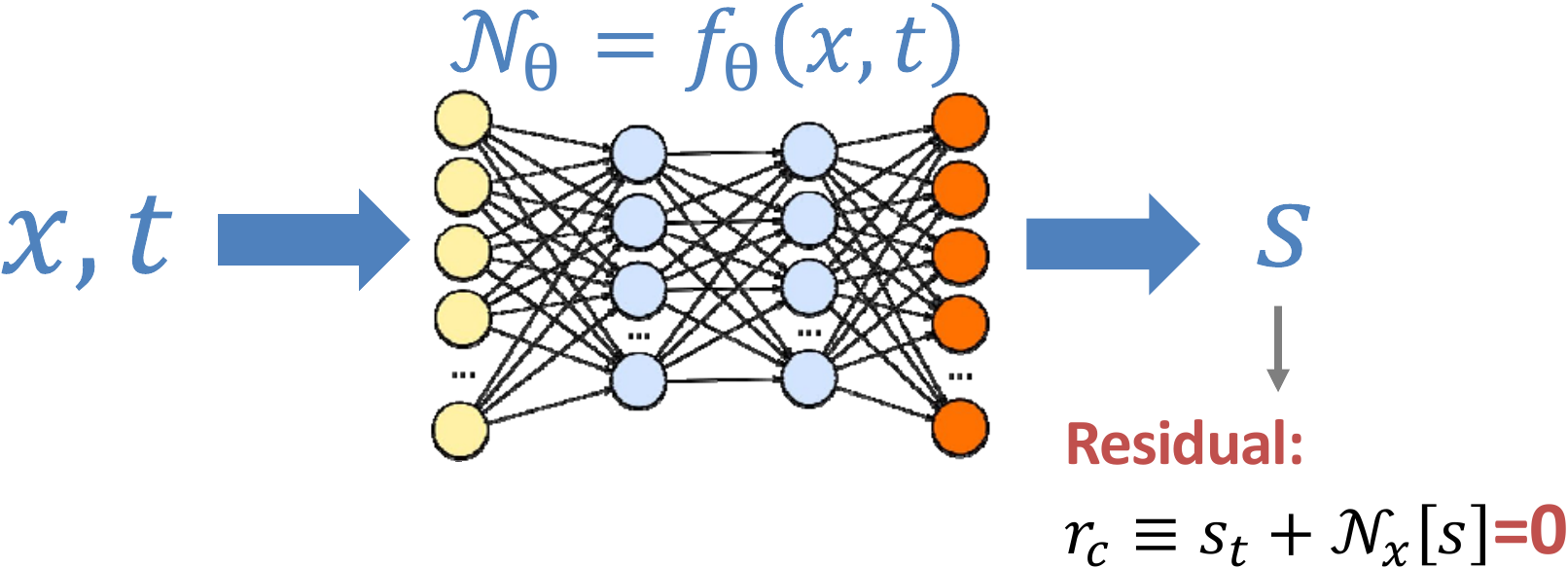}
  \caption{Schematic of PDE solution approximation.}
  \label{fig:pde_sol}
  \vspace{-1.5em}
\end{figure}

One important application of PIDL is to solve generic partial differential equations (PDE). We will briefly introduce the underlying rationale. 
Define a PDE over the ST domain as:
\vspace{-.6cm}
\begin{eqnarray}
\label{equ:PDE_u}
& \mathbf{s}_t(x,t) + \mathcal{N}_x[\mathbf{s}(x,t)] = 0, (x,t)\in {\cal D}, \\
& \small{\mathcal{B}}[\mathbf{s}(x,t)]=0,  \ (x,t) \in   \partial {\cal D}, \nonumber \\
& \small{\mathcal{I}}[\mathbf{s}(x,0)]=0, \nonumber
\vspace{-1em}
\end{eqnarray}
where, 
$\mathcal{N}_x(\cdot)$ is the nonlinear differential operator, 
$\mathcal{B},\mathcal{I}$ are boundary and initial condition operators, respectively, $\partial {\cal D} = \{(0,t)|t\in[0,T]\} \cup \{(L,t)|t\in[0,T]\}$ is the set of ST points on the boundary of the domain ${\cal D}$,
$\mathbf{s}(x,t)$ is the exact solution of the PDE. Now we will approximate the PDE solution, $\mathbf{s}(x,t)$, by a DNN parametrized by $\theta$, $f_{\theta}(x,t)$, which is PUNN. 
If this PUNN is exactly equivalent to the PDE solution, then we have
\begin{equation}
\begin{eqnsize}
\label{equ:ftheta}
    f_{\theta}(x,t) + \mathcal{N}_x[f_{\theta}(x,t)] = 0, (x,t)\in D.
\end{eqnsize}
\end{equation}
Otherwise we define a residual function $r_{c}(x,t) = [f_{\theta}(x,t)]_t + \mathcal{N}_x[f_{\theta}(x,t)]$. 
If PUNN is well trained, the residual needs to be as close to zero as possible. 
Fig.~\ref{fig:pde_sol} describes the schematic of using PUNN to approximate PDE solutions.

Physics are usually encoded as \emph{governing equations}, \emph{physical constrains}, or \emph{regularity terms} in loss functions when training PUNNs \cite{alber2019integrating}. 
When ``soft regularization" is implemented \cite{krishnapriyan2021characterizing}, the loss of PIDL is a weighted sum of two distance measures: 
one for the distance between observed action and predicted one (aka. \emph{data discrepancy})
and the other for the distance between computed action from physics and predicted one (aka. \emph{physics discrepancy}). 
Its specific forms will be detailed in Secs.~\ref{sec:pidl_deterministic}-\ref{sec:pidl_uq}. 
\vspace{-1em}
\subsection{Traffic state estimation}

Traffic state inference is a central problem in transportation engineering and serves as the foundation for traffic operation and management. 
\begin{defn}
\textbf{Traffic state estimation (TSE)}
infers traffic states in single-lane or multi-lane along a stretch of highway or arterial segments over a period of time, represented by traffic density (\emph{veh/km/lane}, denoted by $\rho$), traffic velocity (\emph{km/lane/hour}, denoted by $u$), and traffic flux or volume (\emph{veh/lane/hour}, denoted by $q$) using noisy observations from sparse traffic sensors \cite{Seo-17}. 
The ultimate goal of TSE is traffic management and control building on the inferred traffic states.
\end{defn}
\begin{rem}
\begin{enumerate}
\item 
Three traffic quantities are connected via a universal formula: 
\vspace{-0.5em}
\begin{equation}\label{eq:q}
\begin{eqnsize}
    q=\rho u.\vspace{-0.5em}
\end{eqnsize}
\end{equation}
Knowing two of them automatically derives another. 
Thus, in this paper, we will primarily focus on $\rho, u$, and $q$ can be derived by Eq.~\ref{eq:q}.
\item $q\in [0,q_{max}], \rho\in [0,\rho_{jam}], u\in [0,u_{max}]$, where $q_{max}$ is the capacity of a road, $\rho_{jam}$ is the jam density (i.e., the bumper-to-bumper traffic scenario), and $u_{max}$ is the maximum speed (and usually represented by speed limit). How to calibrate these parameters are shown in Tab.~\ref{tab:model_cal}.
\end{enumerate}
\end{rem}

TSE is essentially end-to-end learning from the ST domain to labels. 
Denote a mapping parameterized by $\theta$ as $f_{\theta}(\cdot)$, from a ST domain $(\mathbf{x},t)\in {\cal D}$ to traffic states $\mathbf{s}=\lbrace \rho, u \rbrace$: 
\vspace{-0.5em}
\begin{equation}\label{eqn: generic_map}
\begin{eqnsize}
f_{\theta}: (\mathbf{x},t) \longrightarrow \mathbf{s}=\lbrace \rho, u \rbrace. \vspace{-0.5em}
\end{eqnsize}
\end{equation}

The key question is to find a set of parameters $\theta^*$ and the functional form $f_{\theta}(\cdot)$ that fit observational data the best. 

\begin{rem}
TSE can be implemented using supervised learning as presented in this paper, where the physics-based models are used to regularize the training of the data-driven models.
TSE can also be formulated as unsupervised learning such as matrix/tensor completion that estimates unknown traffic states \cite{chen2021low,chen2020nonconvex}. Instead of using physics-based models, matrix/tensor completion methods use prior knowledge such as low-rank property to regularize the estimation. We would like to pinpoint here that such prior knowledge regularization can also be integrated into our framework by including the rank of the predicted traffic state matrix in the PIDL loss function. 
\end{rem}

Traffic sensors range from conventional ones placed on roadside infrastructure (in Eulerian coordinate), such as 
inductive loop detectors, 
roadside closed-circuit television (CCTV) or surveillance cameras, 
to in-vehicle sensors (in Lagrangian coordinate), 
including Global Positioning System (GPS), on-board cameras, LiDARs,
and smart phones. 
The emerging traffic sensors mounted on connected and automated vehicles (CAVs) are expected to generate terabytes of streaming data~\cite{cv_sas}, 
which can serve as ``probe vehicles" or ``floating cars" for traffic measurements.
Future cities will be radically transformed by the Internet of Things (IoT), which will provide ubiquitous connectivity between physical infrastructure, mobile assets, humans, and control systems~\cite{IoTLitRev}
via communication networks (e.g., 5G~\cite{IoT5GSur20}, DSRC~\cite{dsrcAndCell,meinrenken2020using}, x-haul fiber networks~\cite{xHaulel19}, edge-cloud~\cite{surveyPP18} and cloud servers). 

\begin{figure}[h!]
\vspace{-1em}
\centering
  \includegraphics[scale=0.35]{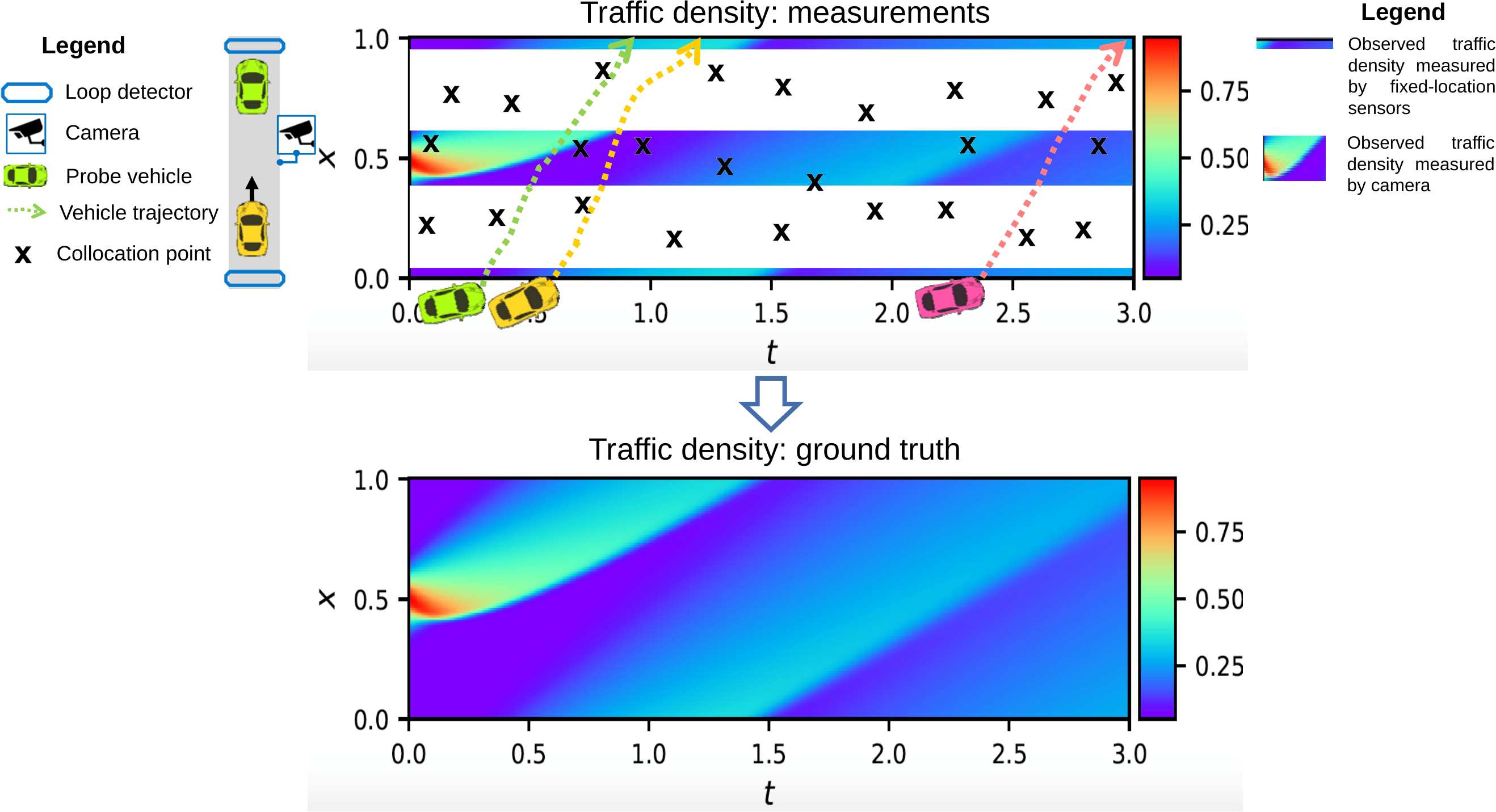}
  \caption{Data types for TSE (adapted from \cite{Seo-17}, including fixed location sensors (in blue hexagon), roadside camera, and collocation points (in black cross)).}
  \label{fig:data_tse}
  \vspace{-1em}
\end{figure}

Fig.~\ref{fig:data_tse} illustrates the observational data types for TSE. 
Building on individual vehicle trajectories, 
we can aggregate velocity and density for each discretized cell and time interval. 
With the availability of high-resolution multi-modality data, we should also consider developing disaggregate methods for TSE that can directly use individual trajectories or images as inputs.

In the TSE problem, 
the physics-based approach refers to the scientific hypotheses about the evolution of traffic flow on micro-, meso-, and macro-scale; 
while the ML approach refers to data-driven models that mimic human intelligence using deep neural networks, reinforcement learning, imitation learning, and other advanced data science methods \cite{di2021survey}.

\subsubsection{Physics-based models}

The field of transportation has been buttressed by rich theories and models tracing back to as early as the 1930s \cite{greenshields1935study}. 
A large amount of theories have since been successfully developed to explain real-world traffic phenomena, to prognose and diagnose anomalies for operation and management, and to make predictions for planning and management. 
    

These models make ample use of scientific knowledge and theories about transportation systems, ranging from closed-form solutions to numerical models and simulations. 
Transportation models have demonstrated their analytical and predictive power in the past few decades. 
For example, microscopic car-following models and macroscopic traffic flow models succeed to capture transient traffic behavior, including shock waves and stop-and-go phenomenon. 

Model based approach relies on traffic flow models for single- or multi-lane, single- or multi-class traffic flow. 
Traffic models include 
the first-order models like Lighthill-Whitham-Richards (LWR)~\cite{Lighthill-1955, Richards-1956}, 
and the second-order models like Payne-Whitham (PW)~\cite{Payne-1971, Whitham-1974} and Aw-Rascle-Zhang (ARZ)~\cite{Aw-2002, zhang-2002}. 

The first constitutive law that needs to satisfy is a conservation law (CL) or transport equation, meaning that inflow equals outflow when there is no source or sink. Mathematically, 
\vspace{-0.5em}
\begin{equation}
\begin{eqnsize}
\label{eq:model_conserv}
(CL) \ \ \ \ \ \ \rho_t + (\rho u)_x=0, \  (x,t)\in {\cal D}. 
\vspace{-0.5em}
\end{eqnsize}
\end{equation}

\begin{figure}
\centering
  \includegraphics[scale=0.4]{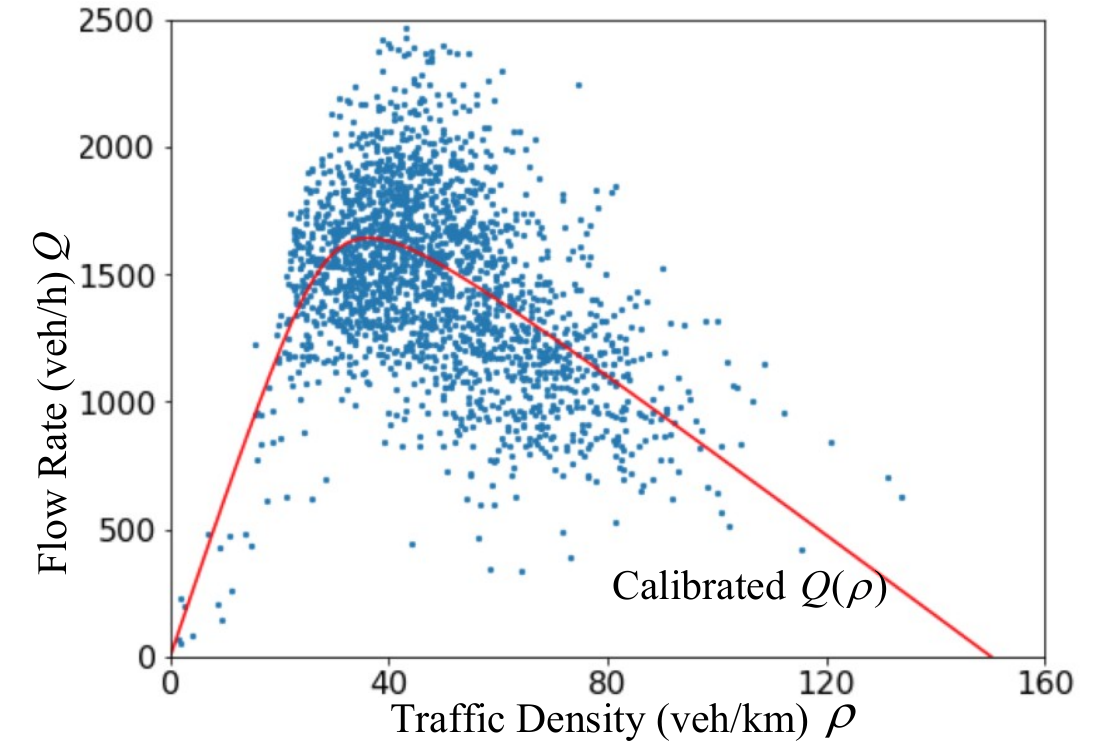}
  \caption{Fundamental diagram (in red line) with data (blue dots).}\vspace{-1em}
  \label{fig:ch5-FD}
  \vspace{-1em}
\end{figure}

A second equation that stipulates the relation between $\rho, u$ can be a fundamental diagram (FD) (for the first-order traffic flow model) or a moment equation (for the second-order traffic flow model): 
\vspace{-0.5em}
\begin{equation}
\begin{eqnsize}
\label{eq:model_fd}
\begin{aligned}
(FD) \ \ \ \
u= U(\rho), \mbox{ (first-order)}  \\
\ \ \ \ \ \ \ \ u_t + uu_x = g(U(\rho)), \mbox{ (second-order)}.
\end{aligned}
\vspace{-0.5em}
\end{eqnsize}
\end{equation}
where $U(\cdot)$ is a fundamental diagram, a mapping from traffic density to velocity, 
and $g(U(\rho))$ is a nonlinear function of $U(\rho)$. 
A fundamental diagram can also be a mapping from density to volume/flux, as exemplified in Fig.~\ref{fig:ch5-FD} calibrated by a real-world traffic dataset. 
In the literature, several FD functions are proposed and interested readers can refer to \cite{turner201175} for a comprehensive survey. 


When a traffic flow model is selected as the underlying dynamical system, data assimilation is employed to  find ``the most likely state" and use observations to correct the model prediction, including extended Kalman filter (EKF)~\cite{Yibing-2008, Yibing-2009, di2010hybrid}, unscented KF~\cite{Mihaylova-2006}, ensemble KF~\cite{Blandin-2012}), 
and particle filter~\cite{Mihaylova-2004}.


To quantify the uncertainty of TSE problems, the model-based approach usually makes a prior assumption about the distribution of traffic states by adding a Brownian motion on the top of deterministic traffic flow models, leading to Gaussian stochastic traffic flow models. 
There is the other school of literature that derives intrinsic stochastic traffic flow models with more complex probabilistic distributions \cite{davis1994estimating,kang1995estimation,di2010hybrid,jabari2012stochastic}. 
With a stochastic traffic flow model, large population approximation or fluid limit is applied to extract the first and second moments of the stochastic processes to facilitate the application of filtering methods.

\subsubsection{Data-driven approach}

The data-driven approach aims to learn traffic dynamics 
directly from data.
Machine learning extracts knowledge, pattern and models, automatically from large volumes of data. 
DL, especially DNN, 
has revived the interest of the scientific community since 2006 \cite{hinton2006fast}. 
Using ML for TSE is primarily focused on data imputation leveraging temporal or spatial correlation, including  
autoregressive integrated moving average~\cite{Zhong-2004}, 
probabilistic graphical models~\cite{Ni-2005},
k-nearest neighbors~\cite{Tak-2016}, 
principal component analysis~\cite{Li-2014, Tan-2014}, 
long short term memory model~\cite{ma2015long}. 
The majority of these methods assume that a traffic quantity at a time interval or within a cell depends on its historical or neighboring values, regardless of the physical characteristics of traffic flow. 
Accordingly, data-driven approach is not as popular as the model-based one and does not achieve as a high accuracy as the latter \cite{Seo-17}. 
More recent ML techniques aim to model non-linearity in traffic dynamics, leveraging deep hidden layers together with sparse autoregressive technique~\cite{Polson-2017}
and fuzzy neural network~\cite{Tang-2020}. 
With the advantage of both model and data based approaches, it is natural to consider a hybrid one in TSE. 

\subsubsection{PIDL}

In the pioneering work~\cite{Raissi-2018a, Raissi-2018b}, PIDL was proposed as an alternative solver of PDEs. 
Since its inception, PIDL has become an increasingly popular tool for data-driven solution or discovery of nonlinear dynamical systems 
in various engineering areas \cite{Yang-2019,Maziar-2019,Fang-2020,Maziar-2020,rai2020driven}. 
While PIDL has increasingly demonstrated its predictive power in various fields, transportation modeling is lagging behind in combining both physics and data aspects. 
\vspace{-1em}
\subsection{Two classes of problems}
\label{subsec:2prob}

The existing literature on PIDL aims to solve two classes of problems: 
(1) PDE solution inference,
and (2) uncertainty quantification. 
In the next two sections, we will detail these two problems one by one. 
\vspace{-1em}
\section{PIDL for deterministic TSE}\label{sec:pidl_deterministic}

\subsection{PIDL for traffic state estimation (PIDL-TSE)}\label{subsec:prob_stat}


\begin{defn}
\textbf{PIDL for traffic state estimation (PIDL-TSE)}
aims to infer the spatiotemporal fields of traffic states by integrating physics based and deep learning methods. 

Define the (labeled) observation set as ${\cal O}_d, {\cal O}_p$, the boundary and initial observation sets as ${\cal B}, {\cal I}$, and the (unlabeled) collocation point set as ${\cal C}$ below:
\vspace{-2em}
\begin{equation}
\begin{eqnsize}
\label{equ-3-x1}
\left\{ \hspace{-0.6em} {\begin{array}{*{20}l}
    \  {\cal O}_s=\{(\mathbf{x}^{(i_s)}, t^{(i_s)});  (\hat{\rho}^{(i_s)},\hat{u}^{(i_s)})\}_{i_s=1}^{N_{o_s}}: \textit{stationary sensors}, \vspace{0.3em} \ \  \\ 
    \  {\cal O}_m=\{\{\mathbf{X}(n,t^{(i_m)})\}_{i_m=1}^{N_{o_m}} \}_{n=1}^{N_n}: \textit{mobile trajectories}, \vspace{0.3em} \ \  \\
    \ {\cal B}=\{t^{(i_b)}; (\hat{\rho}^{(i_b)},\hat{u}^{(i_b)}) \}_{i_b=1}^{N_{b}}: \textit{boundary observation}, \vspace{0.3em} \ \ \\
   \ {\cal I}=\{\mathbf{x}^{(i_0)}; (\hat{\rho}^{(i_0)},\hat{u}^{(i_0)})\}_{i_0=1}^{N_{0}}: \textit{initial observation}, \vspace{0.3em} \ \ \\
   \ {\cal C}=\{(\mathbf{x}^{(j)}, t^{(j)})\}_{j=1}^{N_c}: \textit{collocation points}.
\end{array}} \right.\
\end{eqnsize}
\vspace{-0.5em}
\end{equation}
where, $i_s,i_m$ are the indices of data collected from stationary and mobile sensors, respectively;
$i_b,i_0$ are the indices of data collected from boundary and initial conditions, respectively;
and $j$ is still the index of collocation points. 
The numbers of stationary sensor data, mobile data, boundary and initial conditions, and collocation points are denoted as $N_{o_s},N_{o_m},N_{b},N_{0}, N_c$, respectively.
The number of mobile trajectories is denoted as $N_{n}$. 
$\mathbf{X}(n,t^{(i_m)})$ is the $n^{th}$ vehicle's position at time $t^{(i_m)}$.
Observation data in ${\cal O}$ are limited to the time and locations where traffic sensors are placed. 
In contrast, collocation points ${\cal C}$ have neither measurement requirements nor location limitations, and is thus controllable. 

\end{defn}

In the next two sections,  we will elaborate the PIDL-TSE framework on the architecture of HCG and training methods.
\vspace{-1em}
\subsection{Hybrid computational graph (HCG)}

HCG is a tool we have invented to facilitate the visualization of the two components, namely, PUNN and PICG, and how they are wired. 
Over an HCG, the architecture of the PUNN and the PICG, the loss function to train the PUNN, and the training paradigm can be defined visually. 
A computational graph, establishing mathematical consistency across scales, is a labeled directed graph 
whose nodes are (un)observable physical quantities representing input information, intermediate quantities, and target objectives. 
The directed edges connecting physical quantities represent the dependency of a target variable on source variables,  
carrying a mathematical or ML mapping from a source to a target quantity.  
A path from a source 
to the observable output quantities 
represents one configuration of a model. 
A model configuration is to establish a path within the HCG~\cite{wang2021non}.


\vspace{-1em}
\subsection{Training paradigms}


Once the physics model is selected, we need to determine the sequence of parameter calibration prior to, or during the training of PUNN. 
The former corresponds to solely infer time-dependent traffic flow fields, 
and the latter corresponds to system identification of traffic states \cite{Raissi-2018b}.


\subsubsection{Sequential training}
Sequential training aims to first calibrate parameters of the PICG (i.e., parameter calibration) and then encode the known physics into the PUNN for training. 
Parameter calibration has been extensively studied in TSE using 
non-linear programming \cite{seo2015traffic}, 
genetic algorithm \cite{cremer1981parameter}, 
least-squares fitting \cite{Fan-2013,kurzhanskiy2010active,fan2013comparative,deng2013traffic}, 
and kernel smoothing\cite{Ngoduy-2011}. 
The physics based parameters include $\rho_{max},u_{max}$ and other nominal parameters. 
Tab.~\ref{tab:model_cal} summarizes the existing methods for model discovery.
\begin{table}\centering
\begin{threeparttable}
	\centering
	\caption{State-of-the-art approaches for parameter calibration \label{tab:model_cal}}
	\begin{tabular}{|p{0.1 cm}<{\centering}|p{1.1 cm}<{\centering} |p{2. cm}<{\centering}||p{0.9 cm}<{\centering}|p{0.9 cm}<{\centering} |p{0.9 cm}<{\centering} |}
		\hline
	    \multicolumn{2}{|c|}{\multirow{2}{*}{Method}} & \multirow{2}{*}{Description} & $\rho_{max}$ & $\rho_{critical}$ & $u_{max}$\\ \cline{4-6}
	    \multicolumn{2}{|c|}{}  & & Max. density & Critical density & Max. speed \\ \hline\hline
	   \multirow{3}{*}[-4.5em]{\rotatebox[origin=c]{90}{Sequential training}}  
		&
		Calibrate each parameter separately & Each parameter carries certain physical meaning  
		& segment length divided by ave. vehicle length 
		& traffic density at capacity 
		& speed limit or max. value\\ \cline{2-6} 
	    & Calibrate parameter and predict state jointly & augment states with parameters estimated using DA \cite{davis1994estimating,kang1995estimation,di2010hybrid,jabari2012stochastic}  & \multicolumn{3}{c|}{\parbox[t]{3 cm}{tuning along with other hyperparameters in DNNs}}  \\ \cline{2-6}
	    & Calibrate FD & fit parameters associated with a pre-selected FD \cite{Fan-2013,kurzhanskiy2010active,fan2013comparative,cremer1981parameter,Huang-Jiheng-2020,seo2015traffic,Ngoduy-2011} & density at $u=0$  & density at $\max q$ & velocity at maximum \\ \hline\hline
	    \multirow{2}{*}[-4.5em]{\rotatebox[origin=c]{90}{Joint training}}  
	    & Calibrate FD & fit parameters assiciated with a pre-selected FD along with parameters of DNNs \cite{Barreau-2021-CDC,shi2021physics_arxiv,shi2021aaai} & density at $u=0$ & density at $\max q$ & velocity at maximum \\ 
	    \cline{2-6}
		& ML surrogate & reduce variable and parameter sizes while maintaining the minimum physical relevance \cite{brunton2016discovering,Liu-2021-IFAC-PapersOnLine,shi2021physics} & \multicolumn{3}{c|}{\parbox[t]{3 cm}{parametrized in DNNs}}  \\ \hline 
		\end{tabular}
    \end{threeparttable}
    \vspace{-1em}
\end{table}

Sequential training is the default paradigm in most PIDL-related studies, with a focus on how to make the training robust and stable, when large-scale NNs and complicated physics-informed loss functions are involved. 
A growing amount of works aim to develop more robust NN architectures and training algorithms for PIDL. For example, one can use adaptive activation function by introducing a scalable hyper-parameter in the activation function at some layers of the PUNN to improve the convergence rate and solution accuracy~\cite{Jagtap-2020}. The adaptive activation function has also been used in DeLISA (deep learning based iteration scheme approximation), which adopts the implicit multistep method and Runge-Kutta method for time iteration scheme to construct the learning loss when training the PUNN~\cite{li2022delisa}.  
To perform an efficient and stable convergence in the training phase, \cite{Wang-Sifan-2022} investigates the training dynamics using neural tangent kernel (NTK) theory and proposes a NTK-guided gradient descent algorithm to adaptively adjust the hyperparameters for each loss component. 
New algorithms and computational frameworks for improving general PIDL training are currently a popular research area, and we refer readers to~\cite{karniadakis2021physics} by
Karniadakis \textit{et al.} for a detailed survey on this topic.
    
   


    


\subsubsection{Joint training}

Physics parameters and hyperparameters of the PUNN and the PICG are updated \textit{simultaneously} or \textit{iteratively} in the training process. 
All the existing literature on PIDL-TSE employs simultaneous updating of all parameters associated with both PICG and PUNN altogether, which will be our focus below. 
However, we would like to pinpoint that there are increasing interests in training both modules iteratively \cite{zhang2022physics}, which could be a future direction to improve the training efficiency of PIDL-TSE.

\paragraph{Challenges}

The PUNN in PIDL is a typical deep learning component that most training techniques can apply. 
In contrast, the training challenges incurred by the PICG with unknown physics parameters are nontrivial, and accordingly, substantial research and additional adaptive efforts are in need.

First, some traffic flow models may include a large number of physical parameters that need to discover in TSE, and it is challenging to train all the parameters at once. For example, the three-parameter LWR model in section~\ref{sec:3param_LWR} involves 5 parameters, and it is reported that the direct joint training for all the parameters with real-world noisy data leads to unsatisfactory results~\cite{shi2021physics}. 
For this issue, the alternating direction method of multipliers (ADMM) method~\cite{Boyds-2011} is an option to improve training stability, i.e., to train one subset of physical parameters at a time with the rest fixed. The advanced ADMM variant, deep learning ADMM (dlADMM), may further address the global convergence problems in non-convex optimization with a faster learning efficiency~\cite{Wang-Junxiang-2019}. 

Second, a highly-sophisticated traffic flow model may contain complicated terms that are unfriendly to the differentiation-based learning, making the model parameter discovery less satisfactory for real data. In this case, the structural design of PICG plays an important role to make the framework trainable. Specifically, additional efforts such as variable conversion, decomposition and factorization need to be made before encoding to have the structure learnable and the loss to converge. Alternatively, as will be discussed in section~\ref{sec:ML_surrogate}, one can use an ML surrogate, such as a small NN, to represent the complicated terms in PICG to avoid tackling them directly~\cite{shi2021physics}.


\paragraph{ML surrogate} \label{sec:ML_surrogate}



\begin{figure}[h!]
\centering
  \includegraphics[scale=0.44]{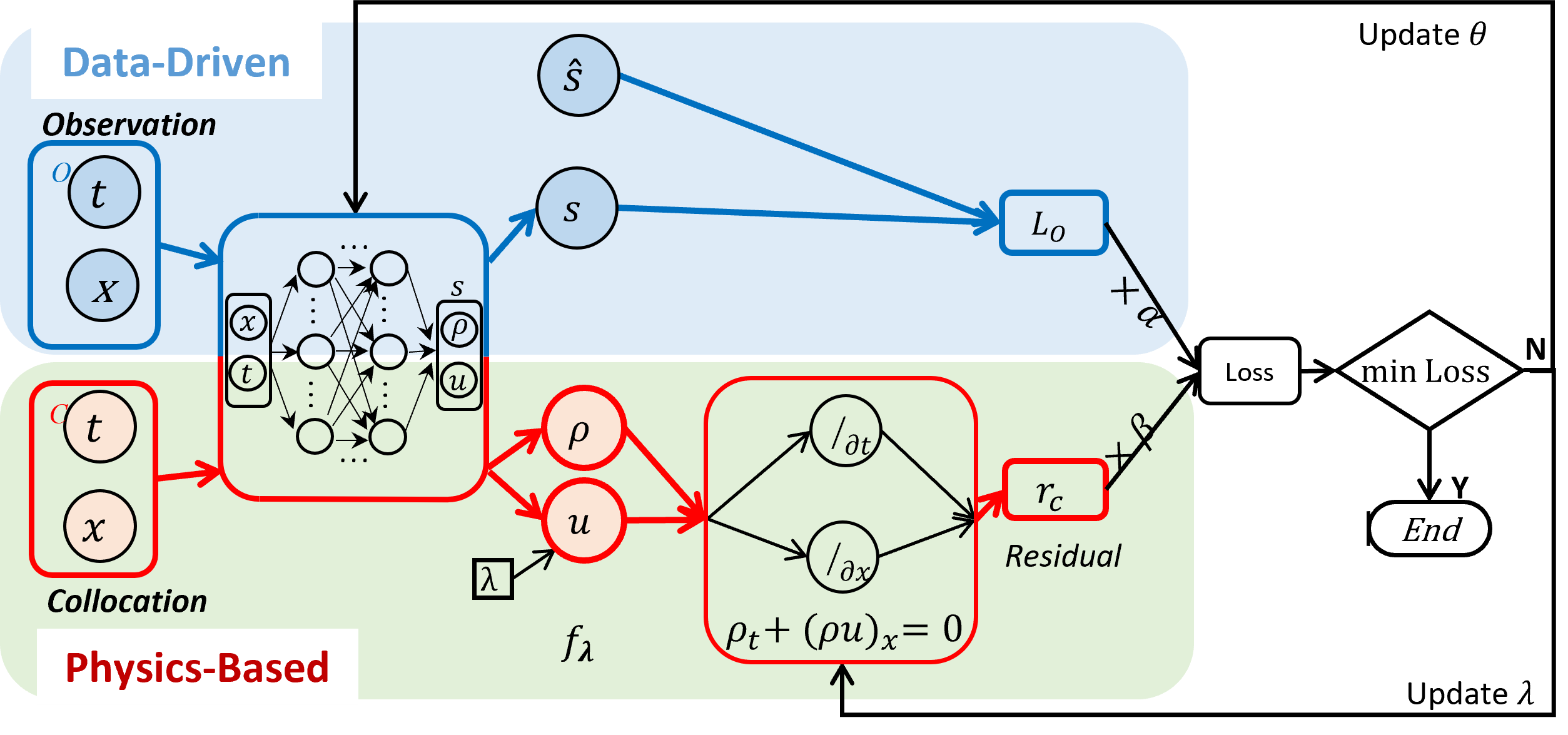}
  \caption{Flowchart of joint training of PIDL.}
  \label{fig:joint_flow}
\end{figure}

When physics and PUNN are jointly trained, Fig.~\ref{fig:joint_flow} further demonstrates the flow of simultaneously tuning parameters in the physics model and the hyperparameters in the PUNN. 
The top blue box encloses the data-driven component that contributes to the loss function, 
and the bottom red box encloses the physics-based component.

Note that in Fig.~\ref{fig:joint_flow}, we omit details about how $\rho$ and $u$ interact within a PUNN. 
These two quantities can be generated in sequence or in parallel. 
When $\rho$ is generated first, $u$ can be derived from an FD that stipulates the relation between $\rho$ and $u$. 
Otherwise, one PUNN can be trained to generate both $\rho, u$ altogether or two PUNNs are trained to generate $\rho$ and $u$ separately. 
Since $\rho, u$ need to satisfy the CL, but generated out of PUNNs cannot guarantee that constraint. 
Thus, $\rho, u$ are then fed into the physics component to impose this constraint on these quantities. 

To inject the minimum amount of knowledge like the universal CL defined in Eq.~\ref{eq:model_conserv} to PIDL, and leave out those based on assumptions like FD defined in Eq.~\ref{eq:model_fd}, 
we propose an ML surrogate model that replaces the mapping from traffic density to velocity as a DNN and learns the relation between these two quantities purely from data.

A fundamental diagram that establishes a relation between $\rho$ and $u$, can be treated as an embedded physics component within traffic flow models.
According to empirical studies, the relation between $\rho$ and $u$ is generally not a one-to-one mapping, especially in the congested regime. 
Thus, it is natural to employ an ML surrogate component to characterize the interaction between these two traffic quantities. 
We can further explore to what extent the addition of surrogates affects the performance and where ML surrogates are appropriate to use.

Tab.~\ref{tab:lit_pidl} summarizes the existing studies that employ PIDL for TSE.  To leverage PIDL for the TSE problem is firstly proposed by Huang \textit{et al.}~\cite{Huang-Jiheng-2020} and Shi \textit{et al.}~\cite{shi2021physics_arxiv, shi2021aaai,shi2021physics}, concurrently and independently.

\begin{table}
\vspace{-1em}
\begin{threeparttable}
	\caption{State-of-the-art PIDL-TSE} \label{tab:lit_pidl}
	\begin{tabular}{|m{0.015 cm} |m{0.4 cm} ||m{0.8 cm}<{\centering} |m{5.3 cm}<{\centering} ||m{0.3 cm}<{\centering}|}
		\hline
		\multicolumn{2}{|c||}{Physics} & Data & Descriptions  & Ref. \\ \hline
		\multirow{4}{*}[-2em]{\rotatebox[origin=c]{90}{First-order model}}  
		
		& LWR & synthetic (Lax-Hopf method) &\makecell[l]{Integrated the Greenshields-based LWR to \\ PIDL and validated it using loop detectors \\ as well as randomly placed sensors}  & \cite{Huang-Jiheng-2020} \\ \cline{2-5}	
		
		& LWR & \makecell[l]{numeri- \\cal, \\NGSIM} &\rule{0pt}{3.0em}\makecell[l]{Presented the use of PIDL to solve \\ Greenshields-based and three-parameter-based \\ LWR models, and demonstrated its advantages \\ using the real-world dataset}  & \cite{shi2021physics_arxiv} \\  
		
		\cline{2-5}	
		
		& LWR & \makecell[l]{numeri- \\cal} & \makecell[l]{Studied the general partial-state reconstruction \\ problem for traffic flow estimation, and used \\ PIDL encoded with LWR to counterbalance \\ the small number of probe vehicle data}   & \cite{Barreau-2021-CLDC}
		\\ \cline{2-5}

		& \makecell{FL1,\\LWR} & SUMO simulation & \rule{0pt}{0em} \makecell[l]{Integrated a coupled micro-macro model, \\ combining the Follow-the-Leader (FL1) model \\ and LWR model, to PIDL for TSE, which can \\ use the velocity info from probe vehicles}   & \cite{Liu-2021-IFAC-PapersOnLine, Barreau-2021-CDC}
		\\ \hline\hline
		\multirow{2}{*}[1.5em]{\rotatebox[origin=c]{90}{Second-order  model}}  
		& \multirow{2}{*}{\makecell[l]{LWR\\ \& \\ ARZ}} & \multirow{2}{*}{\makecell[l]{ \makecell[l]{numeri- \\cal, \\NGSIM}}}
		 & \rule{0pt}{0em} \makecell[l]{Applied the PIDL-based TSE to the second-\\ order ARZ with observations from both loop \\ detectors and probe vehicles, and estimate both \\ $\rho$ and $u$ in parallel}  & \cite{shi2021aaai}  \\
		\cline{4-5}
		& & & \rule{0pt}{0em} \makecell[l]{ Proposed the idea of integrating ML surrogate \\ (e.g., an NN) into the physics-based component \\ in PICG to represent the complicated FD \\ relation. Improved estimation accuracy achieved  \\and unknown FD relation is learned}  & \cite{shi2021physics} \\ 
		\hline
		\end{tabular}
    \end{threeparttable}
    \vspace{-0.5em}
\end{table}

Next, we will demonstrate how to design the architecture of PIDL and the corresponding PICG.
We will first present a numerical example to demonstrate how the physics law of three-parameter-based LWR is injected into the PICG to inform the training of the PUNNs, and then compare all the existing architectures on a real-world dataset. 
\vspace{-1em}
\subsection{Numerical data validation for three-parameter-based LWR}\label{sec:3param_LWR}



In this example, we show the ability of PIDL for the traffic dynamics governed by the three-parameter-based LWR traffic flow model on a ring road. Mathematically,
\vspace{-1em}

\begin{equation}
\begin{eqnsize}
\label{equ-3-B-1}
\rho_t + (Q(\rho))_x=\epsilon \rho_{xx},  \  x\in [0,1], \  t\in [0,3], 
\end{eqnsize}
\end{equation}

\noindent
where  $\epsilon = 0.005 $. 
The initial and boundary conditions are $\rho(x,0)=0.1+0.8e^{-25(x-0.5)^2}$ and $\rho(0,t)=\rho(1,t)$, respectively. 

In this model, three-parameter flux function~\cite{Fan-2013} is employed: $Q(\rho)=\rho U(\rho)=\sigma (a+(b-a)\frac{\rho}{\rho_{max}}-\sqrt{1+y^2})$, where $a=\sqrt{1+(\delta p)^2}$, $b=\sqrt{1+(\delta (1-p))^2}$ and $y = \delta (\frac{\rho}{\rho_{max}}-p)$. In the model, $\delta$, $p$ and $\sigma$ are the three free parameters as the function is named. The parameters $\sigma$ and $p$ control the maximum flow rate and critical density (where the flow is maximized), respectively. $\delta$ controls the roundness level of $Q(\rho)$. In addition to the above-mentioned three parameters, we also have $\rho_{max}$ and diffusion coefficient $\epsilon$ as part of the model parameters. In this numerical example, we set  $\delta = 5$, $p=2$, $\sigma = 1$, $\rho_{max}=1$ and $\epsilon = 0.005$.

Given the bell-shaped initial density condition, we apply the Godunov scheme to solve Eqs.~\ref{equ-3-B-1} on 240 (space) $\times$ 960 (time) grid points evenly deployed throughout the $[0,1]\times [0,3]$  domain.  


\begin{figure}[H]
\centering
  \includegraphics[scale=0.48]{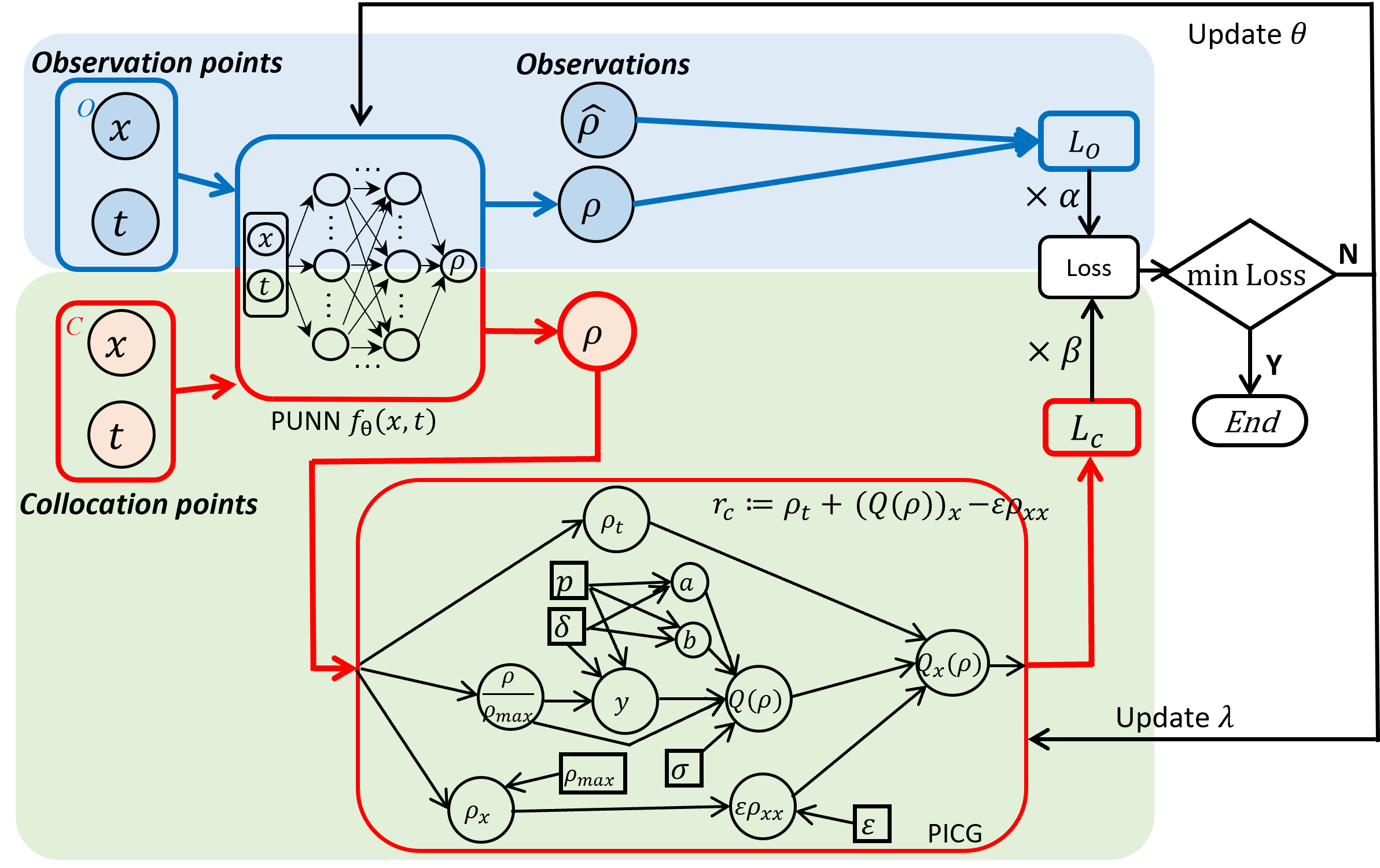}
  \caption{PIDL flowchart for three-parameter-based LWR, consisting of a PUNN for traffic density estimation and a PICG for calculating the residual, where $\lambda = (\delta,p,\sigma,\rho_{max},\epsilon)$.}
  \label{fig:ch3-3_param_PINN}
  \vspace{-1.5em}
\end{figure}

The PIDL architecture that encodes the LWR model is shown in Fig.~\ref{fig:ch3-3_param_PINN}. This architecture consists of a PUNN for traffic density estimation, followed by a PICG for calculating the residual $r_c:=\rho_t + (Q(\rho))_x-\epsilon \rho_{xx}$ on collocation points. 
The estimated traffic density $\rho$ is calculated by the PUNN $f_{\theta}(x,t)$, which is an NN and maps a spatiotemporal point $(x,t)$ directly to $\rho$, i.e., $\rho=f_{\theta}(x,t)$. 
PUNN $f_{\theta}(x,t)$, parameterized by $\theta$, is designed as a fully-connected feedforward neural network with 8 hidden layers and 20 hidden nodes in each hidden layer. Hyperbolic tangent function (tanh)  is used as the activation function for hidden neurons. By replacing $\rho$ with $f_{\theta}$, we have $r_c:=[f_{\theta}]_t + (Q(f_{\theta}))_x-\epsilon [f_{\theta}]_{xx}$ in this case. With the estimated $\rho$ and the observations $\hat{\rho}$ on the observation points,  we can obtain the data loss $L_o$. In contrast, in PICG, connecting weights are fixed and the activation function of each node is designed to conduct specific nonlinear operation for calculating an intermediate value of~$r_c$. The physics discrepancy $L_c$ is the mean square of $r_c$ on collocation points.

To customize the training of PIDL to  Eqs.~\ref{equ-3-B-1}, we  need to additionally introduce \textit{ boundary collocation points} ${\cal C}_B=\{(0,t^{(i_b)})| i_b = 1,...,N_b\} \cup \{(1,t^{(i_b)})| i_b = 1,...,N_b\}$, for learning the two boundary conditions. Different from the $\mathcal{B}$ in Eqs.~\ref{equ-3-x1}, observations on boundary points are not required here in ${\cal C}_B$. Then, we obtain the following loss:




\vspace{-1em}
\vspace{-0.1em}
\begin{equation}
\begin{eqnsize}
\label{equ-3-B-2}
Loss_{\theta} = \alpha \cdot L_o + \beta  \cdot L_c + \gamma  \cdot L_b,
\vspace{-0.7em}
\end{eqnsize}
\end{equation}
\begin{equation*}
\begin{eqnsize}
    \begin{aligned}
     \text{where,} \ \ L_o= \frac{\alpha}{N_o} \sum_{i=1}^{N_o} |f_{\theta}(x^{(i)}, t^{(i)})-\hat{\rho}^{(i)}|^2 \ \ (data\ loss), \\[-0.5em]
     L_c = \frac{\beta}{N_c}\sum_{j=1}^{N_c} |r_c(x^{(j)},t^{(j)})|^2 \ \ (physics\ loss), \\[-1em]
     L_b = \frac{\gamma}{N_b} \sum_{i_b=1}^{N_b} |f_{\theta}(0,t^{(i_b)})-f_{\theta}(1,t^{(i_b)})|^2\ \ (boundary\ loss). \\
    \end{aligned}
    \vspace{-0.5em}
\end{eqnsize}
\end{equation*}

\noindent
Note that because $r_c:=[f_{\theta}]_t + (Q(f_{\theta}))_x-\epsilon [f_{\theta}]_{xx}$, $r_c$ is affected by $\theta$. Also, boundary collocation points are used to calculate the boundary loss $L_b$. Because $L_b$ might change with different scenarios, it is ignored from Fig.~\ref{fig:ch3-3_param_PINN} for simplicity.

\textbf{TSE and system identification using loop detectors}: 
In this experiment, five model variables $\delta$, $p$, $\sigma$, $\rho_{max}$, and $\epsilon$ are encoded as learning variables in PICG depicted in Fig.~\ref{fig:ch3-3_param_PINN}. 
Define $\lambda = (\delta,p,\sigma,\rho_{max},\epsilon)$, and the residual $r_c$ is affected by both $\theta$ and $\lambda$, resulting in the objective $Loss_{\theta, \lambda}$. We now use observations from loop detectors, i.e., only the traffic density at certain locations where loop detectors are installed can be observed. By default, loop detectors are evenly located along the road.

\begin{figure}[h]
\centering
  \includegraphics[width=1.0 \columnwidth]{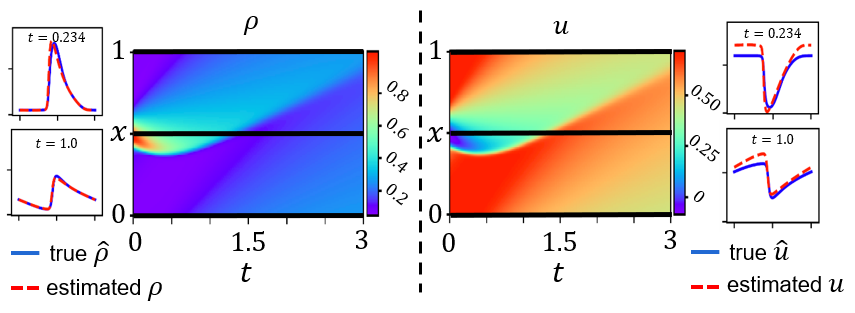}
  \caption{Estimated traffic density $\rho$ (left half) and traffic velocity $u$ (right half) of the PIDL when the number of loop detectors is 3, where the horizontal black lines in the heatmap represent the sensor positions. In each half, the prediction heatmap and snapshots at certain time points are presented. Note that the PUNN does not predict $u$ directly, and instead, it is calculated by $Q(\rho)/\rho$ in the post-processing.} \vspace{-1.2em}
  \label{fig:LWR-3-Param-snapshot}
\end{figure}

We use $N_c = 150,000$ collocation points and other experimental configurations are given in SUPPLEMENTARY~A1. 
We conduct PIDL-based TSE experiments using different numbers of loop detectors to solve $(\theta^*, \lambda^*) = \mathrm{argmin}_{\theta,\lambda}\  Loss_{\theta, \lambda}$. In addition to the traffic density estimation errors of $\rho(x,t;\theta^*)$, we evaluate the estimated model parameters $\lambda^*$ using the $\mathbb{L}^2$ relative error (RE) and present them in percentage. Fig.~\ref{fig:LWR-3-Param-snapshot} presents a visualization of the estimated traffic density $\rho$ (left half) and traffic velocity (right half) of the PIDL. The comparision at certain time points are presented. Note that the PUNN in Fig.~\ref{fig:ch3-3_param_PINN} does not predict $u$ directly, and instead, it is calculated by $Q(\rho)/\rho$ in the post-processing.

More results are provided in 
SUPPLEMENTARY~Table~I, 
 and the observation is that the PIDL architecture in Fig.~\ref{fig:ch3-3_param_PINN} with five loop detectors can achieve a satisfactory performance on both traffic density estimation and system identification. In general, more loop detectors can help our model improve the TSE accuracy, as well as the convergence to the true model parameters. Specifically, for five loop detectors, an estimation error of $3.186 \times 10^{-2}$ is obtained, and the model parameters converge to $\delta^* = 4.86236$, $p^* = 0.19193$, $\sigma^*=0.10697$, $\rho^*_{max}=1.00295$ and $\epsilon^*=0.00515$, which are decently close to the ground truth. The observations demonstrate that PIDL can handle both TSE and system identification with 5 loop detectors for the traffic dynamics governed by the three-parameter-based LWR. 

We conduct sensitivity analysis on different numbers of collocation points and how they are identified. The details are presented in SUPPLEMENTARY~Table~II. 
A larger collocation rate (i.e.,  the ratio of the number of collocation points to the number of grid points) is beneficial for both TSE and system identification, because it could make estimation on the collocation points physically consistent by imposing more constraints on the learning process. Empirically, more collocation points can cause longer training time and the performance does not improve too much when a certain collocation rate is reached.

\vspace{-0.5em}
\subsection{Real-world data validation}\label{sec:data}





\begin{figure}[h!]
\centering
  \includegraphics[scale=0.45]{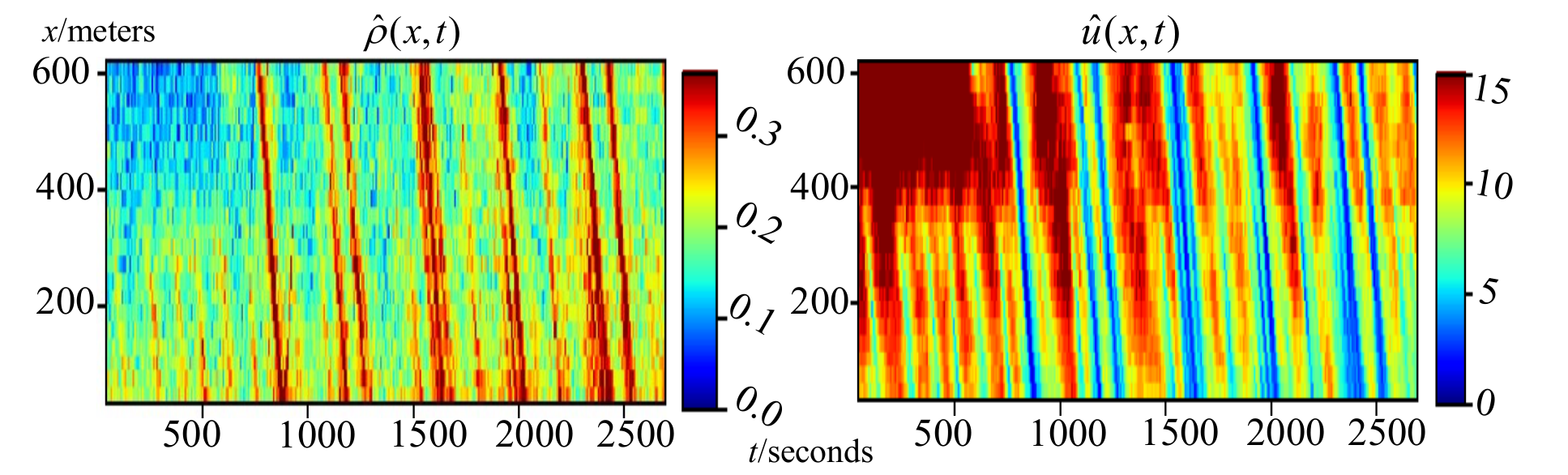}
  \caption{Average traffic density and speed on US 101 highway. Heatmap for the traffic density (left) and velocity (right).}
  \label{fig:ch6-exact}
\end{figure}

It would be interesting to see the performance of state-of-the-art methods based on either physics or data-driven approaches in order to better quantify the added value of the proposed class of approaches
We will use a widely used real-world open dataset, the Next Generation SIMulation (NGSIM) dataset, detailed in Tab.~\ref{tab:data_ngsim}. 
Fig.~\ref{fig:ch6-exact} plots the traffic density heatmap using data collected from US 101 highway.

\begin{figure}[h!]
\vspace{-1.2em}
   \centering
   \subfloat[][RE of the traffic density]{\includegraphics[width=0.5\columnwidth,]{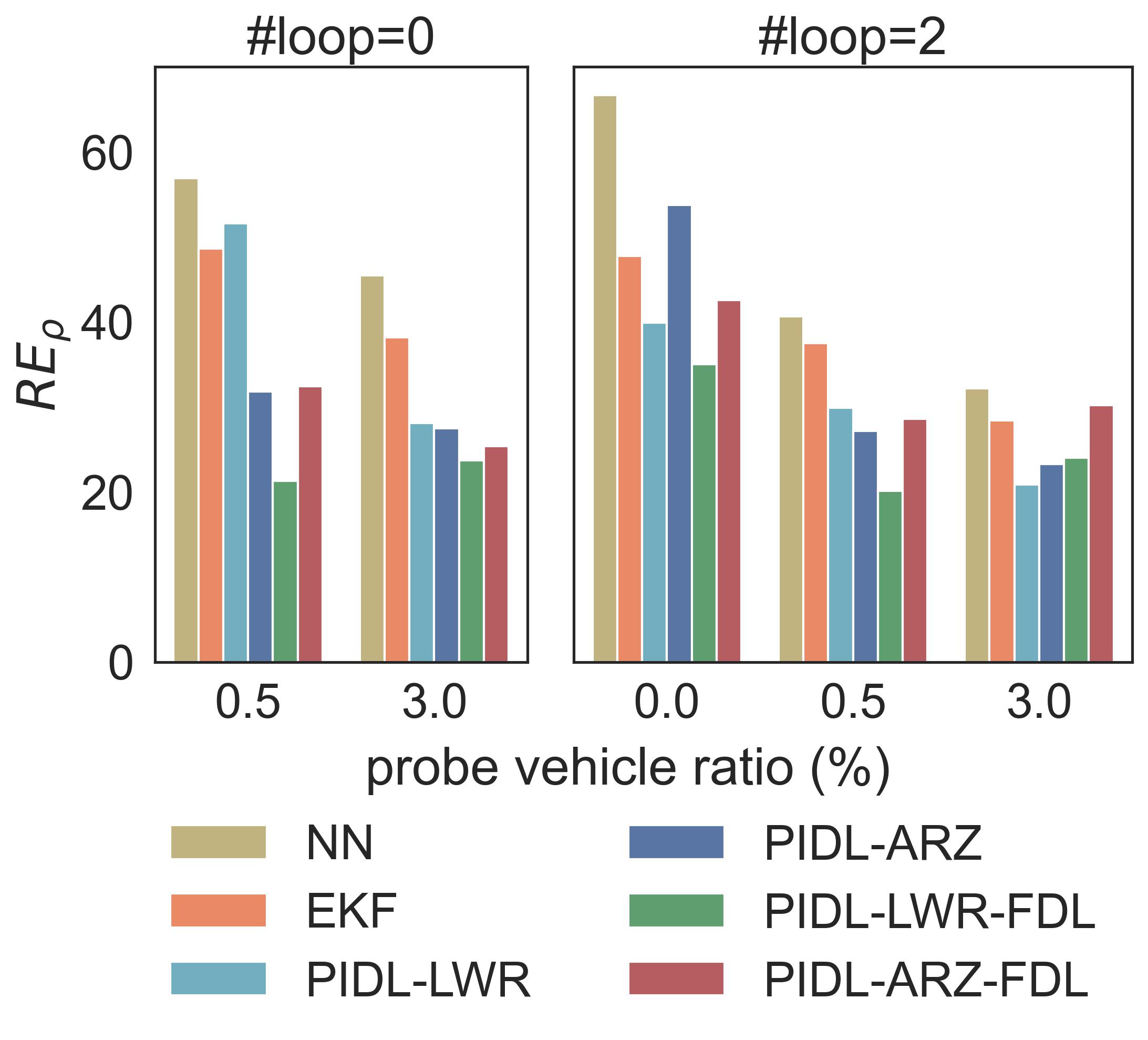}}  
   \subfloat[][RE of the traffic velocity]{\includegraphics[width=0.5\columnwidth]{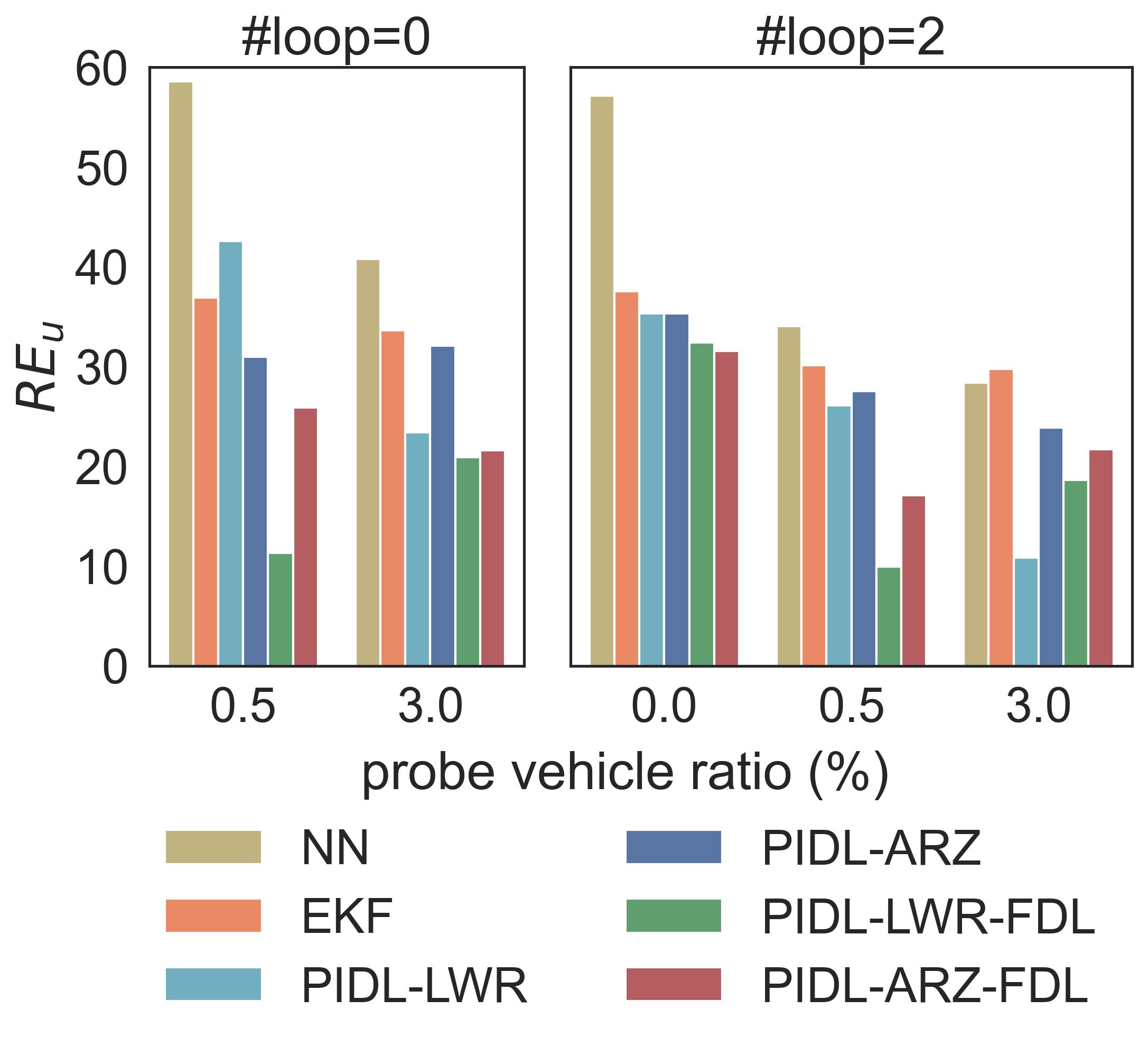}} 
   \caption{Results of the deterministic PIDL models for the NGSIM dataset.}
   \label{fig:results:one_step_all}
   \vspace{-1.5em}
\end{figure}

The performance of 2 baseline models and 4 PIDL variants for deterministic TSE is presented in Fig.~\ref{fig:results:one_step_all}. As shown in the y-axis, the REs of the traffic density and velocity are used for evaluation. The comparison is made under representative combinations of probe vehicle ratios (see x-axis) and numbers of loop detectors (see the title of sub-figures).
We implement an EKF and a pure NN model as the representative pure data-driven and physics-driven baseline approaches, respectively. The EKF makes use of the three parameter-based LWR as the core physics when conducting estimation. The NN only contains the PUNN component in Fig.~\ref{fig:ch3-3_param_PINN}, and uses the first term in Eq.~\ref{equ-3-B-2} as the training loss.
Among the PIDL variants, the PIDL-LWR and PIDL-ARZ are the PIDL models that encodes three parameter-based LWR and Greenshields-based ARZ, respectively, into the PICG. PIDL-LWR-FDL and PIDL-ARZ-FDL are the variant models of PIDL-LWR and PIDL-ARZ by replacing the FD components in the PICG with an embedded neural network (i.e., the FD learner). Note, the FD leaner technique is the one introduced by \cite{shi2021physics}.

\vspace{-0.5em}
\begin{table}[H]\centering
\begin{threeparttable}
	\centering
	\caption{Real-world data description. \label{tab:data_ngsim}}
	\begin{tabular}{|m{.8 cm}<{\centering} ||p{1 cm}<{\centering} |p{1.8 cm}<{\centering} |p{.7 cm}<{\centering} |p{.9 cm}<{\centering} |p{.4 cm}<{\centering}|}
		\hline
		Site & Location & Date & Length (m) & Sampling rate (s) & Lane \# \\ \hline\hline 
		US101\footnote{www.fhwa.dot.gov/publications/research/operations/07030/index.cfm.} & LA, CA & 6/15/2005 & 640 & .1 & 5 \\ \hline
		
		\end{tabular}
		\begin{tablenotes}
      \footnotesize  
      \item Note. Vehicular information includes the position, velocity, acceleration, occupied lane and vehicle class. Time periods include 7:50-8:05 a.m., 8:05-8:20 a.m., 8:20-8:35 a.m.
      We average traffic states across all lanes. 
    \end{tablenotes}
    \end{threeparttable}
\end{table}
\vspace{-1em}

\vspace{-0.5em}

\textbf{Performance of baseline models}: From the experimental results, it is observed that the performance of all models improves as the data amounts increase.
The EKF method performs better than the NN method, especially when the number of observations is small. The results are reasonable because EKF is a physics-driven approach, making sufficient use of the traffic flow model to appropriately estimate unobserved values when limited data are available. However, the model cannot fully capture the complicated traffic dynamics in the real world and the performance improves slowly as the data amount increases. The NN is able to catch up with the EKF when the data is relatively large (see the case with loop=2 and ratio=3.0$\%$).  However, its data
efficiency is low and large amounts of data are needed for accurate TSE.

\textbf{Comparison between PIDL-based models}: The PIDL-based approaches generally outperform the baseline models. PIDL-ARZ achieves lower errors than PIDL-LWR, because that ARZ model is a second-order model which can capture more complicated traffic dynamics and inform the PIDL in a more sophisticated manner. 

\textbf{Effects of using FDL}: Comparing the models with FD learner (PIDL-LWR-FDL and PIDL-ARZ-FDL) to the ones without (PIDL-LWR and PIDL-ARZ), the former generally shows better data efficiency. In PIDL-LWR-FDL and PIDL-ARZ-FDL, the FD equation is replaced by an internal small neural network to learn the hidden FD relation of the real traffic dynamics.
A proper integration of the internal neural network may avoid directly encoding the complicated terms in PIDL and trade off between the sophistication of the model-driven aspect of PIDL and the training flexibility, making the framework a better fit to the TSE problem.  

\textbf{Comparison between PIDL-FDL based models}: PIDL-LWR-FDL can achieve lower errors than PIDL-ARZ-FDL, implying that sophisticated traffic models may not always lead to a better  performance, because the model may contain complicated terms that makes the TSE performance sensitive to the PIDL structural design. With the NGSIM data, PIDL-LWR-FDL can balance the trade-off between the sophistication of PIDL and the training flexibility more properly.

\begin{figure}[h]
\centering
  \includegraphics[scale=0.3]{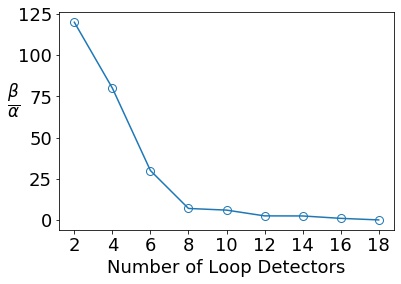}
  \caption{Ratios of the contributions made by the physics-based component and the data-driven component to the optimal performance of PINN.} 
  \label{fig:beta-alpha-deter}
  \vspace{-1.5em}
\end{figure}

 \textbf{Transition from pure physics-driven to data-driven TSE models}: The contributions of the physics-driven and data-driven components can be controlled by tuning the hyperparameters $\alpha$ and $\beta$ in Eqs.~\ref{equ-3-B-2}. Fig.~\ref{fig:beta-alpha-deter} shows how the optimal $\beta/\alpha$ ratio changes as the data size increases. The x-axis is the number of loop detectors, which represents the training data size. The y-axis is the optimal $\beta/\alpha$ corresponding to the minimally achievable estimation errors of the PIDL-LWR methods shown in Fig.~\ref{fig:results:one_step_all}. 
The property of tuning hyperparameters enables the PIDL-based methods to make a smooth transition from a pure physics-driven to pure data-driven TSE model: in the sufficient data regime, by using a small $\beta/\alpha$ ratio, the PIDL performs more like a pure data-driven TSE model to make an ample use of the traffic data and mitigate the issue that the real dynamics cannot be easily modeled by some simple PDEs; while in the ``small'' data regime, by using a large ratio, the PIDL behaves like a pure physics-driven model to generalize better to unobserved domains. 

\section{PIDL for UQ}\label{sec:pidl_uq}

It is a widely regarded modeling and computational challenge 
to quantify how uncertainty 
propagates within dynamical systems that could result in cascading errors, unreliable predictions, and worst of all, non-optimal operation and management strategies. 
It is thus crucial to characterize uncertainties in traffic state estimators and consequently, in traffic management that relies on TSE predictors.
UQ for TSE using PIDL is still at its nascent stage.

\begin{defn}
\textbf{Uncertainty quantification (UQ)}
aims to assess robustness of the developed  model and bound prediction errors of dynamical systems, 
by estimating the probability density of quantities 
with input features and boundary conditions \cite{smith2013uncertainty}. 
Stochastic effects and uncertainties potentially arise from various sources, including variability and errors in measurements, dynamic actions of various entities, model biases, and discretization and algorithmic errors. 
In summary, there are two types of uncertainty  \cite{smith2013uncertainty,abdar2021review} in the context of TSE: 
\begin{enumerate}
    \item Aleatoric uncertainty (or data uncertainty): an endogenous property of data and is thus irreducible, coming from measurement noise, incomplete data, mismatch between training and test data.
    \item Epistemic uncertainty (or knowledge uncertainty, systematic uncertainty, model discrepancy): a property of model arising from inadequate knowledge of the traffic states.
    For example, traffic normally constitutes multi-class vehicles (e.g., passenger cars, motorcycles, and commercial vehicles). 
A single model can lead to insufficiency in capturing diversely manifested behaviors.
\end{enumerate}
\end{defn}

UQ is ``as old as the disciplines of probability and statistics" \cite{smith2013uncertainty}.  
In recent years, its explosive growth in large-scale applications has been bolstered by the advance of big data and new computational models and architectures. 
The conventional UQ techniques include but not limited to: 
sensitivity analysis and robust optimization
\cite{bertsimas2011theory};
probabilistic ensemble methods and Monte-Carlo methods with
multi-level variants \cite{giles2008multilevel,sun2018asymptotically,efendiev2006preconditioning};
stochastic spectral methods \cite{smith2013uncertainty};
and methods based on the Frobenius-Perron and Koopman operators \cite{brunton2016koopman,dietrich2020koopman} for dynamic systems. 

Epistemic uncertainty arising from model discrepancy, often biased, can be compensated by improved domain knowledge, which has received increasing attention, especially with the advance of PIDL.
In the computational architecture of PIDL, the data component can be treated as a compensation term for inaccurate or biased physics supplied by the physics-based component. 
Thus, it is natural to generalize PIDL to UQ, 
where the physics-based component provides partial domain knowledge when stochasticity is propagated within highly nonlinear models, 
and the data-driven component learns extra randomness arising from both data and model errors. 

\begin{defn}
\textbf{UQ for traffic state estimation (UQ-TSE)} aims to capture randomness of traffic states $\hat{\mathbf{s}}=\lbrace \hat{\rho}, \hat{u} \rbrace$ by probabilistic models. It is assumed that $\hat{\mathbf{s}}$ follows the observational distribution, i.e., $\hat{\mathbf{s}} \sim p_{\text{data}}(\hat{\mathbf{s}}|x,t)$. The goal of the UQ-TSE problem is to train a probabilistic model $G_{\theta}$ parameterized by $\theta$ such that the distribution of the prediction $\mathbf{s} \sim p_{\theta}(\mathbf{s}|x,t)$ resembles the distribution of the observation $\hat{\mathbf{s}} \sim p_{\text{data}}(\hat{\mathbf{s}}|x,t)$. One widely-used metric to quantify the discrepancy between $p_{\text{data}}$ and $p_{\theta}$ is the reverse Kullback-Leibler (KL) divergence. 
\end{defn}

Since the majority of literature on UQ-PIDL employs deep generative models, including generative adversarial networks (GAN) \cite{goodfellow2016nips}, normalizing flow \cite{dinh2016density}, and variational autoencoder (VAE) \cite{kingma2013auto}, here we will focus on how to leverage deep generative models for UQ problems. 
Among them, physics-informed generative adversarial network (PhysGAN) is the most widely used model, which has been applied to solve stochastic differential equations \cite{yang2020physics,yang2019adversarial,daw2021pid} and to quantify uncertainty in various domains \cite{siddani2021machine,shi2021physics}. Little has been done on using physics-informed VAE for the UQ-TSE problem, which can be a future direction.

\vspace{-1em}
\subsection{PIDL-UQ for TSE}\label{sec:nn_gnn}

\subsubsection{Physics-informed generative adversarial network (PhysGAN)}

One way to formulate the UQ-TSE problem is to use the generative adversarial network (GAN) \cite{goodfellow2016nips}, 
which imitates the data distribution without specifying an explicit density distribution and can overcome the computational challenge of non-Gaussian likelihood \cite{Yang-2019}, as opposed to using Gaussian process \cite{bilionis2012multi,bajaj2021robust}.

\begin{figure}
   \centering
   \subfloat[][PhysGAN architecture]{ \includegraphics[width=1.0 \columnwidth]{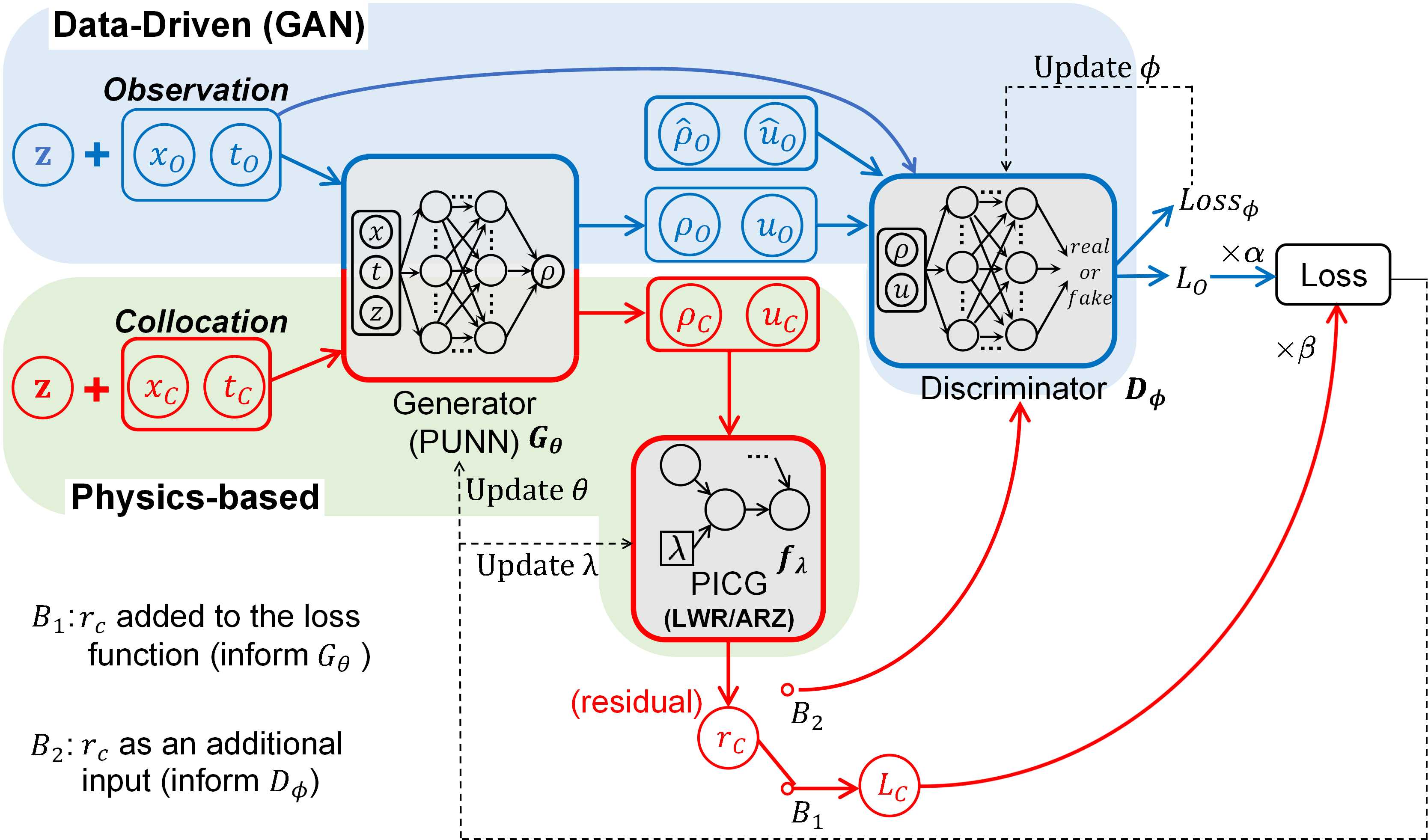}} 
   \\
   \subfloat[][PhysFlow architecture]{ \includegraphics[width=0.65 \columnwidth]{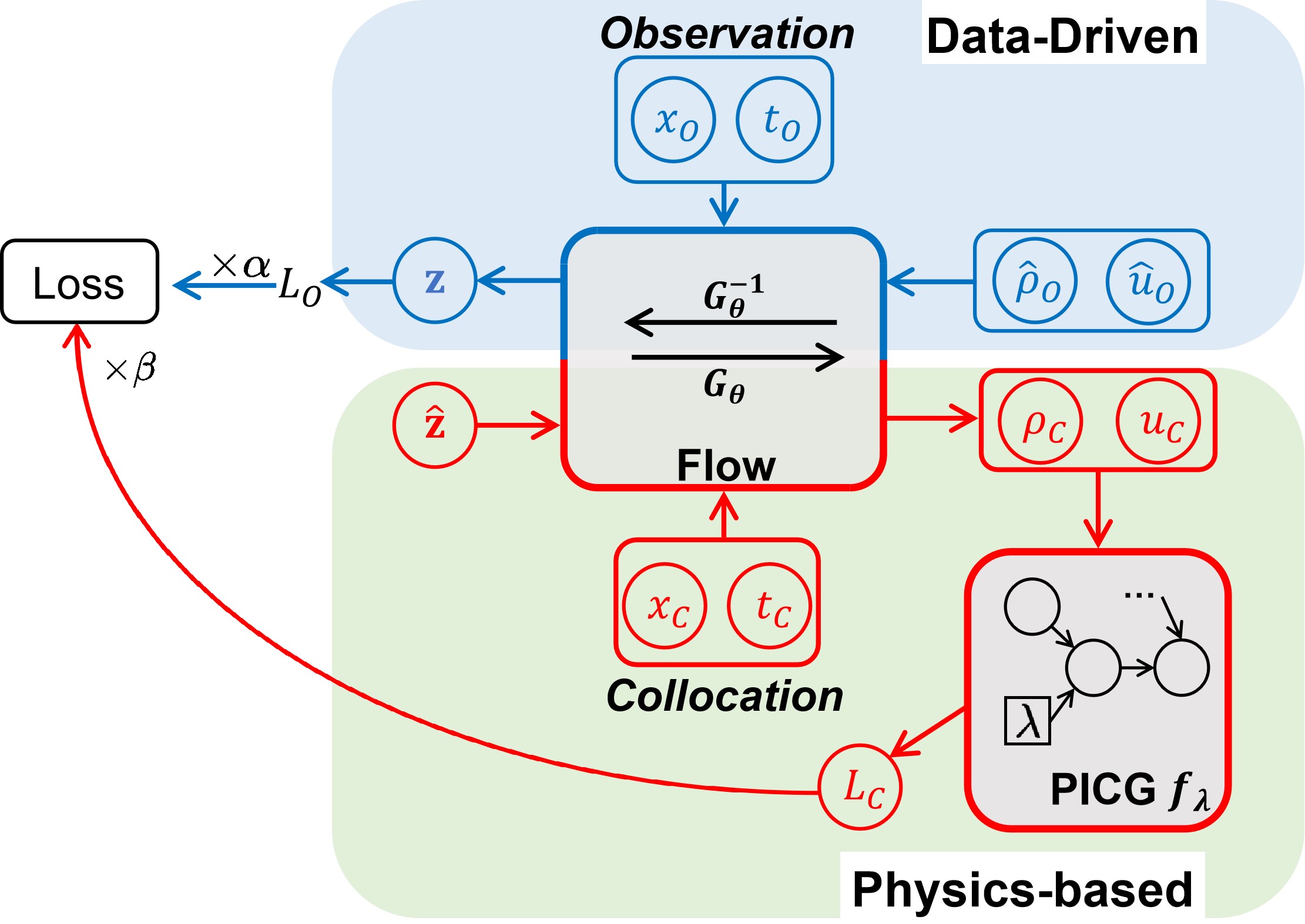}} 
   \\
   \subfloat[][PhysFlowGAN architecture]{ \includegraphics[width=1.0 \columnwidth]{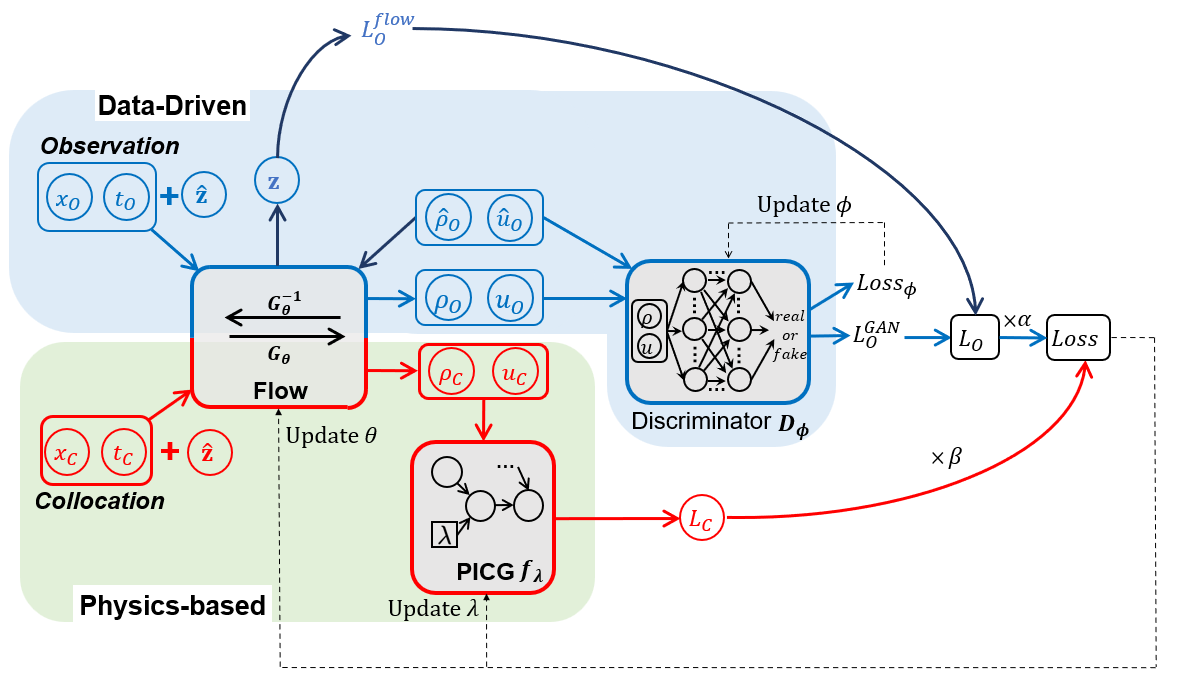}}  
   \caption{PIDL architecture for UQ-TSE. The PhysGAN (a) consists of a generator, a discriminator, and a PICG. The PhysFlow (b) consists of a normalizing flow and a PICG. The PhysFlowGAN (c) consists of a normalizing flow, a discriminator, and a PICG.}
   \label{fig:physgan_structure}
\end{figure}

Now we will formulate the UQ problem in the context of conditional GANs. 
The generator $G_{\theta}$ learns the mapping from the input $(x,t)$ and a random noise $z$ to the traffic state $\mathbf{s}$, $G_{\theta}:(x,t,z) \rightarrow \mathbf{s}$, where $\theta$ is the parameter of the generator. The objective of the generator $G_{\theta}$ is to fool an adversarially trained discriminator $D_{\phi}: (x,t,\mathbf{s}) \rightarrow [0,1]$. 
The loss functions of the GAN are depicted as below:
\begin{equation}
\begin{eqnsize}
\label{equ-arz}
\begin{aligned}
&\hspace{-4.2em}Loss_{\theta} \ \ (generator\ loss)\\ &\hspace{-4.2em}= \mathbb{E}_{x,t,\mathbf{z}} \left[ D_{\phi}(x,t, \mathbf{s})\right]
 \simeq \frac{1}{N_o} \sum_{i=1}^{N_o}D_{\phi}(x^{(i)},t^{(i)}, \mathbf{s}^{(i)}),
\end{aligned}
\end{eqnsize}
\end{equation}
\begin{equation}
\begin{eqnsize}
\label{equ-arz}
\begin{aligned}
&Loss_{\phi} \ \ (discriminator\ loss)\\ &= -\mathbb{E}_{x,t,\mathbf{z}} \left[ \ln D_{\phi}(x,t, \mathbf{s})\right]-\mathbb{E}_{x,t,\hat{\mathbf{s}}}\left[\ln (1-D_{\phi}(x,t,\hat{\mathbf{s}}))\right],\\
& \simeq -\frac{1}{N_o} \hspace{-0.2em} \sum_{i=1}^{N_o}
     \ln D_{\phi}(x^{(i)},t_o^{(i)}, \mathbf{s}^{(i)}) \hspace{-0.1em} + \hspace{-0.1em}\ln(1-D_{\phi}(x^{(i)},t^{(i)}, \hat{\mathbf{s}}^{(i)})),
\end{aligned}
\end{eqnsize}
\end{equation}
where $\mathbf{s}^{(i)}=G_{\theta}(x^{(i)},t^{(i)},z^{(i)})$ is the predicted traffic state, and $\hat{\mathbf{s}}$ is the ground-truth. 
With physics imposed, the generator loss carries the same form as Eq.~\ref{equ-3-B-2}, and the data loss $L_o$ and boundary loss $L_b$ become:
\vspace{-0.5em}
\begin{equation}
\begin{eqnsize}
    \begin{aligned}
     L_o & = \frac{1}{N_o} \sum\limits_{i=1}^{N_o} D_{\phi}(x^{(i)},t^{(i)}, \mathbf{s}^{(i)}),\\
     L_b &= \frac{1}{N_b} \sum\limits_{i_b=1}^{N_b}  D_{\phi}(x^{(i_b)},t^{(i_b)}, \mathbf{s}^{(i_b)}).
    \end{aligned}
    \label{enq:physgan}
\end{eqnsize}
\vspace{-0.5em}
\end{equation} 
Different PhysGAN variants adopt different ways of integrating physics into GANs, and the exact form of $L_c$ changes accordingly. Below we will introduce the general structure of the PhysGAN and its four variants.\\
\indent The general structure of the PhysGAN is illustrated in Fig.~\ref{fig:physgan_structure}. The top blue box encloses the data-driven component, which is a GAN model consisting of a generator $G_{\theta}$ and a discriminator $D_{\phi}$. Here we omit details about how $\rho$ and $u$ interact within the generator. These two quantities can be generated sharing the same NN or from separate NNs. The PICG can be encoded with either the LWR or the ARZ equations. \\
\indent \textbf{PI-GAN} \cite{yang2019adversarial,zhang2019quantifying,yang2019highly,yang2020physics} calculates the physics loss $L_c$ based on the residual $r_c$, as illustrated in branch $B_1$ of Fig.~\ref{fig:physgan_structure} (a). $L_c$ is added into the loss of the PUNN $L_o$ using the weighted sum. This model is the first and most widely used to encode physics into the generator. \\
\indent \textbf{PID-GAN} \cite{daw2021pid} feeds the residual $r_c$ into the discriminator $D_{\phi}$ to provide additional information on whether the predictions deviate from the physics equations, which helps the discriminator to distinguish between the prediction and the ground-truth. This way of integrating physics is illustrated in branch $B_2$ of Fig.~\ref{fig:physgan_structure} (a). It is worth mentioning that the PID-GAN and the PI-GAN share the same structure of the data-driven component. They differ in how the physics are incorporated, i.e., informing the generator (branch $B_1$) or the discriminator (branch $B_2$). \cite{daw2021pid} shows that, by informing the discriminator, PID-GAN can mitigate the gradient imbalance issue of the PI-GAN.
\\ \indent The above-mentioned two PhysGAN variants use deterministic physics, that is, the parameters in the physics equations are deterministic. \\  
\indent \textbf{Mean-GAN} \cite{mo2022quantifying} incorporates stochastic differential equations into the physics components (illustrated as the PICG in Fig.~\ref{fig:physgan_structure} (a)), where physics parameters are assumed to follow Gaussian distributions. 
The randomness in physics parameters is the source of the epistemic uncertainty, which leads to the randomness in the residual $r_c$. The physics loss is calculated based on the square error of the mean of $r_c$, i.e., $|\frac{1}{N_k}\sum_{i=1}^{N_k} r_c|^2$, where $N_k$ is the number of physics parameter samples. Then the physics loss is included to the loss function of the PUNN using the weighted sum.\\
\indent Within the PICG in Fig.~\ref{fig:physgan_structure} (a), we can also replace the parametric FD with ML surrogates, which is used in the \textbf{PI-GAN-FDL}\cite{yang2019adversarial,mo2022trafficflowgan}.



\subsubsection{Physics-informed normalizing flow (PhysFlow)}

\textbf{PhysFlow} \cite{guo2022normalizing} employs the normalizing flow as an alternative generative model to the GAN. The normalizing flow explicitly estimates the likelihood, and is thus more straightforward to train compared to the GAN model. It estimates the likelihood by constructing an invertible function $G_{\theta}$ that transforms a Gaussian prior $\hat{z}$ to the traffic states $\mathbf{s}$. The structure of the PhysFlow, i.e., PI-Flow, is illustrated in Fig.~\ref{fig:physgan_structure} (b). The top blue box encloses the data-driven component consisting of a normalizing flow model. The inverse function $G_{\theta}^{-1}$ takes as input the traffic states and outputs a predicted prior $z$. The training objective is to make $z$ follow a Gaussian distribution, which can be achieved by the maximum likelihood estimation. The bottom red box encloses the physics component, which is the same as the PI-GAN.

\subsubsection{Physics-informed flow based GAN (PhysFlowGAN) }
\textbf{PhysFlowGAN} combines the merits of GAN, normalizing flow, and PIDL. 
It uses normalizing flow as the generator for explicit likelihood estimation, while exploiting adversarial training with the discriminator to ensure sample quality.  The structure of PhysFlowGAN is shown in Fig.~\ref{fig:physgan_structure} (c), which consists of a normalizing flow, a discriminator, and a PICG. The data loss $L_o$ is composed of of two parts, i.e., $L_o^{GAN}$ that is calculated from the discriminator and $L_o^{flow} $ that is calculated from the normalizing flow. The physics loss is calculated in the same way as PI-GAN. One PhyFlowGAN model, TrafficFlowGAN \cite{mo2022trafficflowgan}, has been applied to the UQ-TSE problem. 

Tab.~\ref{tab:arch_uq} summarizes the hybrid architecture used for the UQ-TSE. 
\begin{table}
\vspace{-1em}
\begin{threeparttable}
	\caption{Architecture used for the UQ-TSE. \label{tab:arch_uq}}
	\begin{tabular}{|m{0.02 cm} |m{0.5 cm} ||m{4.0 cm}<{\centering} |m{1.2 cm}<{\centering} ||m{0.6 cm}<{\centering}|}
		\hline
		\multicolumn{2}{|c||}{Model} & Descriptions & Pros & Ref. \\ \hline
		\multirow{4}{*}[-7em]{\rotatebox[origin=c]{90}{PhysGAN}}  
		
		& PI-GAN &\makecell[l]{$L_c\hspace{-0.3em}=\hspace{-0.3em}\frac{1}{N_c}\sum_{i=1}^{N_c}|r_c|^2$ is added to\\the generator loss function using\\the weighted sum}  & the most widely used  & \cite{yang2019adversarial,zhang2019quantifying,yang2019highly,yang2020physics} \\ \cline{2-5}	
		
		& PID-GAN &\rule{0pt}{3.0em}\makecell[l]{Residual is fed into the discrimina-\\tor:  $D_{\phi}(x^{(j)},t^{(j)}, \mathbf{s}^{(j)}, e^{-|r_c^{(j)}|^2})$, \\which is then averaged over colloc-\\ation points to calculate $L_c$} & can mitigate the gradient imbalance issue  & \cite{daw2021pid} \\  
		
		\cline{2-5}	
		
		
		& Mean-GAN & \makecell[l]{Residual is averaged over the\\physics parameters $\lambda$:\\$L_c\hspace{-0.3em}=\hspace{-0.3em}\frac{1}{N_c}\sum_{i=1}^{N_c}|\frac{1}{N_k}\sum_{i=1}^{N_k}r_{c}|^2$}  & can encode stochastic physics model & \cite{mo2022quantifying}
		\\ \cline{2-5}
		
		& \makecell{PI-\\GAN\\-FDL} & \rule{0pt}{0em} \makecell[l]{The $\rho-u$ relation is approximated\\by ML surrogates;\\physics loss $L_c$ is the same as\\PI-GAN} & requires minimal physics information  & \cite{yang2019adversarial,mo2022trafficflowgan}
		\\ \hline\hline
		\multirow{4}{*}[1.5em]{\rotatebox[origin=c]{90}{PhysFlow}}  
		& PI-Flow & \rule{0pt}{0em} \makecell[l]{The normalizing flow model is\\used as the generator for explicit\\computation of the likelihood;\\physics loss $L_c$ is the same as\\PI-GAN} &  simple structure; easy to train & \cite{guo2022normalizing} 
		\\ \hline\hline
		\multirow{4}{*}[1.5em]{\rotatebox[origin=c]{90}{PhysFlowGAN}}  
		& \makecell{Traffic\\Flow\\GAN} & \rule{0pt}{0em} 
		\makecell[l]{
		The normalizing flow model is
		\\used as the generator for explicit
		\\computation of the likelihood;
		\\ The convolutional neural network
		\\is used as the discriminator to
		\\ensure high sample quality
		\\physics loss $L_c$ is the same as
		\\
		PI-GAN} &  combine the merits of PhysGAN and PhysFlow & \cite{mo2022trafficflowgan} \\ 
		\hline
		\end{tabular}
    \end{threeparttable}
    \vspace{-0.5em}
\end{table}

\subsection{Numerical data validation for Greenshields-based ARZ}
As PI-GAN is the most widely used UQ-TSE model, we conduct numerical experiments using PI-GAN for demonstration. In the next subsection, we will use real-world data to compare the performance of the aforementioned UQ-TSE models.

The ARZ numerical data is generated from Greenshields-based ARZ traffic flow model on a ring road:
\begin{equation}
\begin{eqnsize}
\label{equ-arz}
\begin{aligned}
\rho_t + (Q(\rho))_x=0,  \  x\in [0,1], \  t\in [0,3],\\
\partial_t (u+h(\rho)) + u \cdot \partial_x (u+h(\rho))  = (U_{eq}(\rho) - u)/\tau,\\
h(\rho)=U_{eq}(0)- U_{eq}(\rho). 
\end{aligned}
\end{eqnsize}
\end{equation}
$U_{eq}=u_{max}(1-\rho/\rho_{max})$ is the Greenshields speed function, where $\rho_{max}=1.13$ and $u_{max}=1.02$; $\tau$ is the relaxation time, which is set to 0.02. The boundary condition is $\rho(0,t)=\rho(1,t)$. The initial conditions of $\rho$ and $u$ are $\rho(x,0)=0.1+0.8e^{-25(x-0.5)^2}$ and $u(x,0) = 0.5$, respectively. Gaussian noise following a distribution of $\mathcal{N}(0,0.02)$ is added to impose randomness. The results of applying PI-GAN to ARZ numerical data a
re shown in SUPPLEMENTARY~Table~IV.  Fig.~\ref{fig:ARZ-physflow} illustrates the predicted traffic density (left half) and velocity (right half) of the PI-GAN when the number of loop detectors equals to 3. The snapshots at sampled time steps show a good agreement between the prediction and the ground-truth. 

\begin{figure}[h]
\centering
  \includegraphics[width=1.0 \columnwidth]{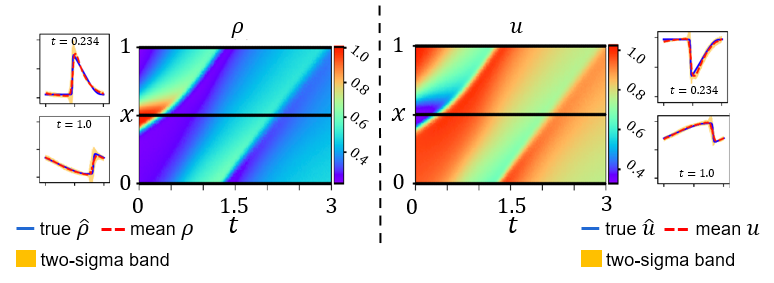}
  \caption{Estimated traffic density $\rho$ (left half) and traffic velocity (right half) of the \tcck{PI-GAN} when the number of loop detectors equal to 3, where the horizontal black lines in the heatmap represent the positions of the loop detectors. In each half, the prediction heatmap and snapshots at certain time points are presented.} \vspace{-1.2em}
  \label{fig:ARZ-physflow}
\end{figure}

\vspace{-1em}
\subsection{Real-world data validation}\label{sec:data}
\begin{figure}
   \centering
   \subfloat[][RE of the traffic density]{\includegraphics[width=0.5\columnwidth,]{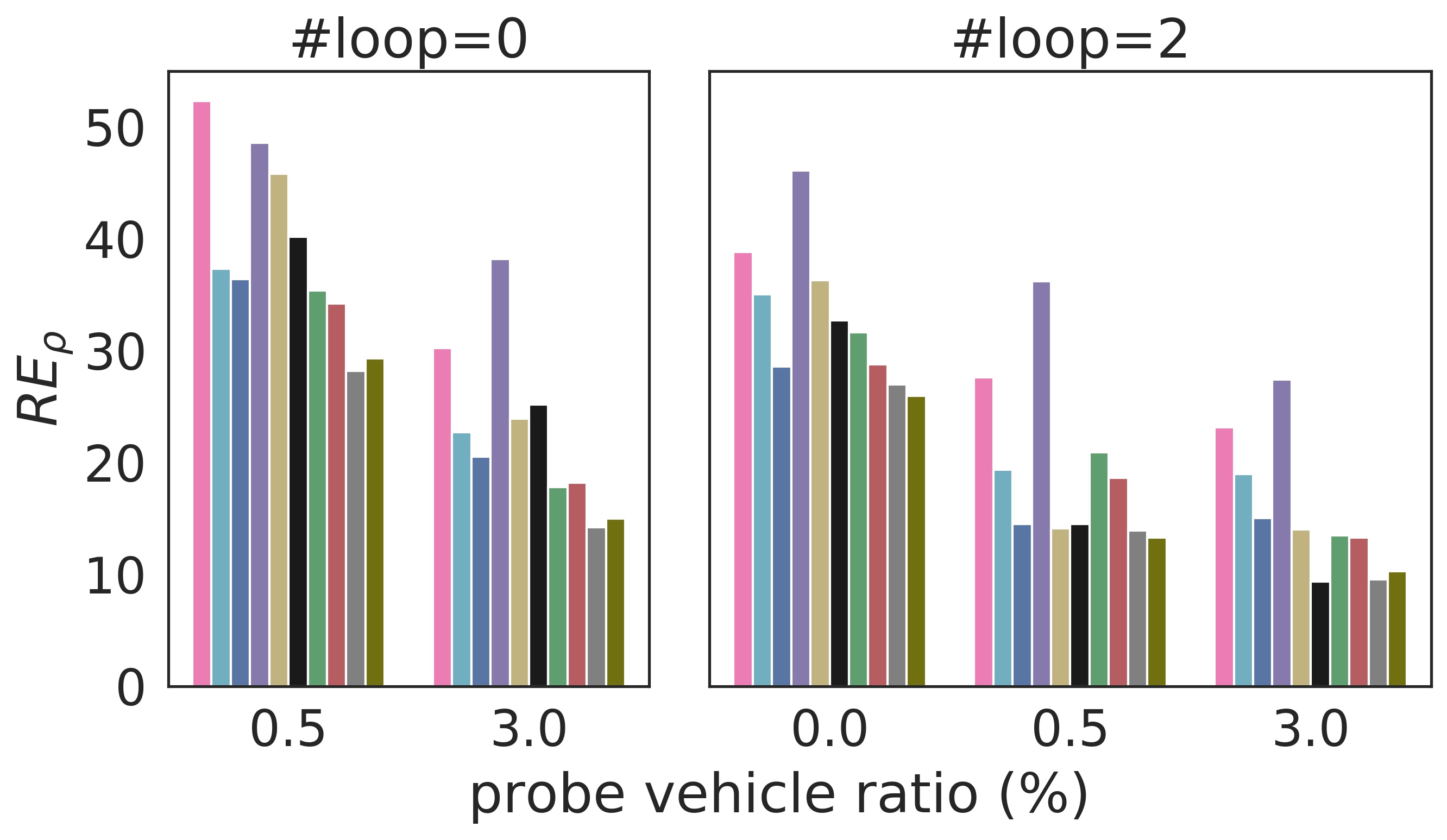}} 
   \subfloat[][RE of the traffic velocity]{\includegraphics[width=0.5\columnwidth]{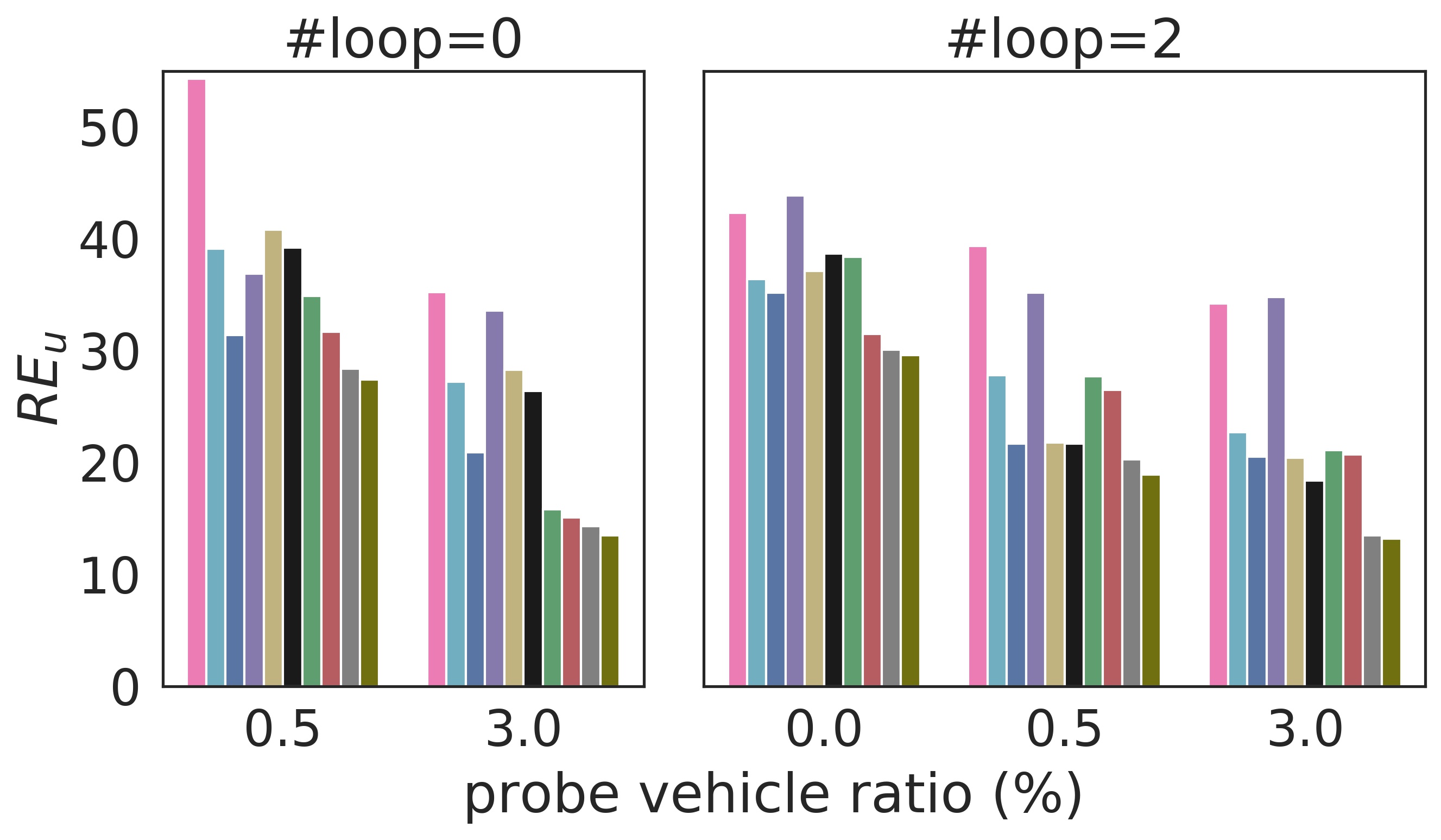}} 
   \\
   \subfloat[][KL of the traffic density]{\includegraphics[width=0.5\columnwidth,]{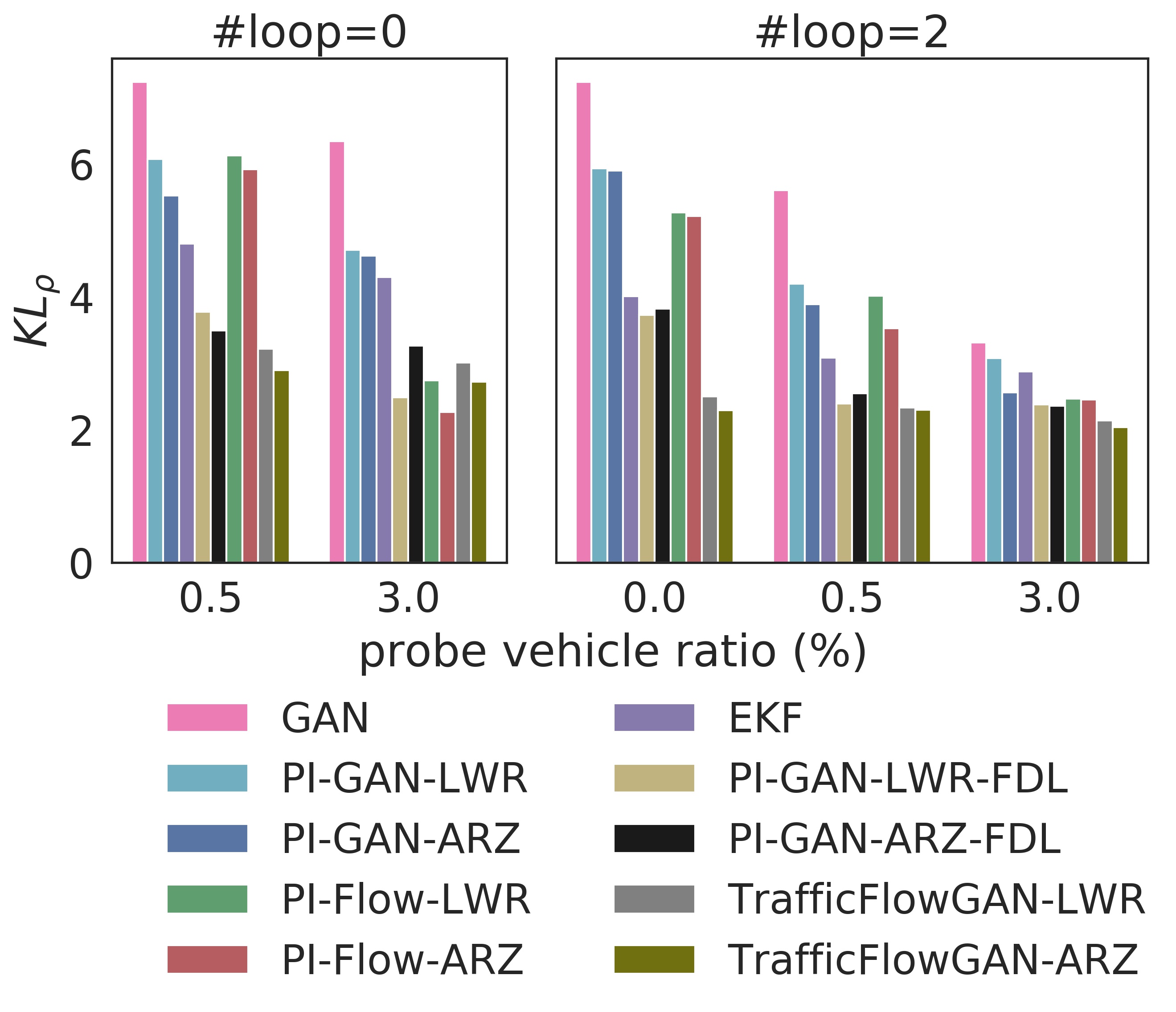}}  
   \subfloat[][KL of the traffic velocity]{\includegraphics[width=0.5\columnwidth]{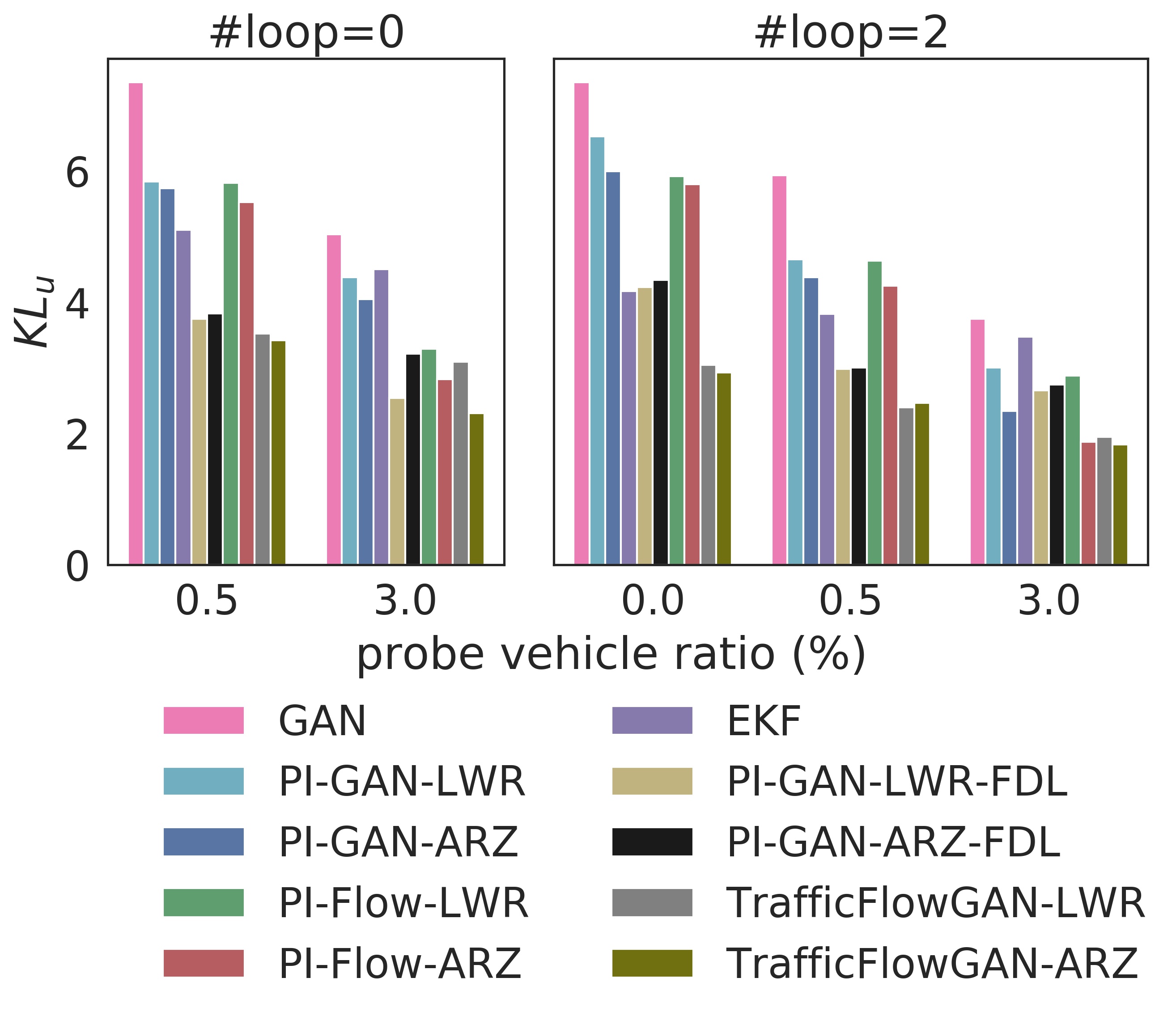}}\\
   \subfloat[][Summary of RE and KL of the trainffic density and velocity of all data sizes]{\includegraphics[width=1\columnwidth]{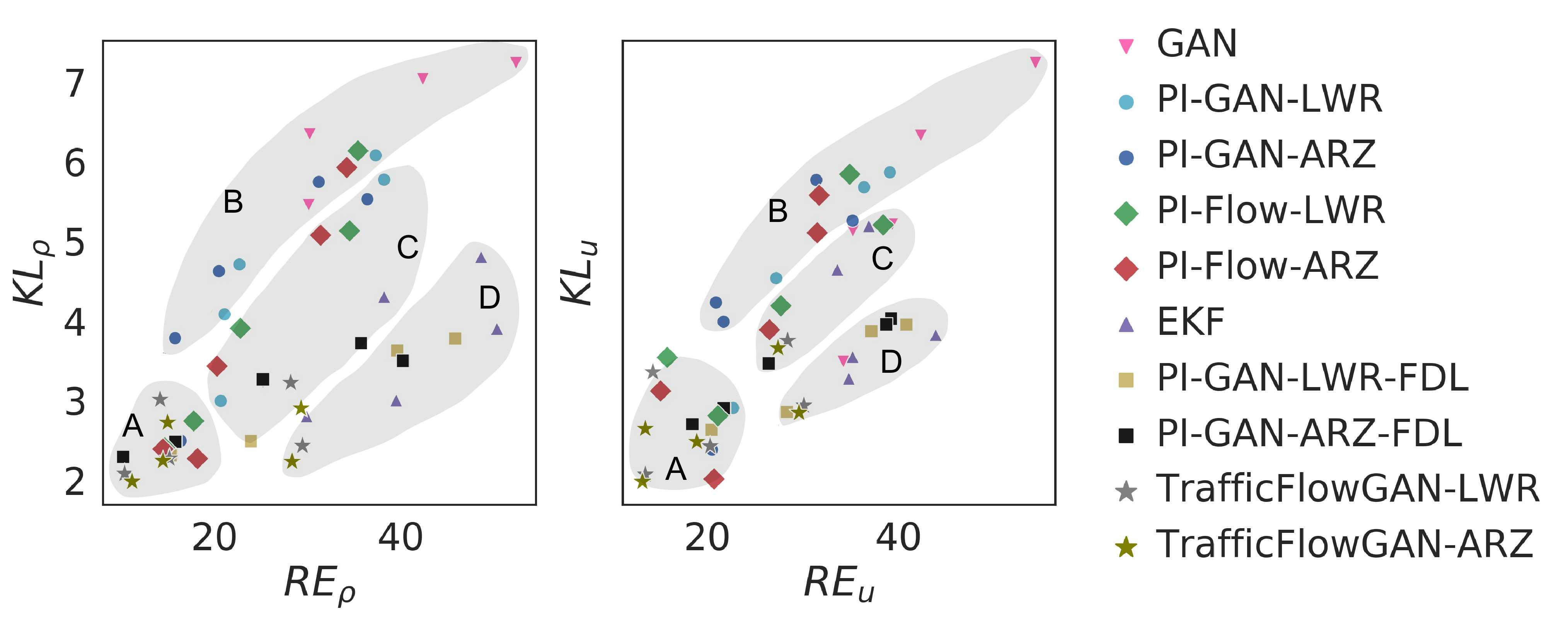}}
   \caption{Results of the PIDL-UQ models for the NGSIM dataset.}
   \label{fig:results:ngsim_sto}
   \vspace{-1em}
\end{figure}

We apply PIDL-UQ models to the NGSIM dataset. 
We use 4 model types, i.e., PI-GAN, PI-GAN-FDL, PI-Flow, and TrafficFlowGAN for demonstration. Each model can be informed by either LWR or ARZ equations, resulting in 8 model variants in total. EKF and GAN are used as baselines. The EKF uses the three parameter-based LWR as the physics. The results are shown in Fig.~\ref{fig:results:ngsim_sto}. As shown in the y-axis, the upper panels are the REs of the traffic density and velocity, and the lower panels are the \tcck{Kullback–Leibler divergence (KL)} of the traffic density and velocity. The comparison is made under representative combinations of probe vehicle ratios (see x-axis) and numbers of loop detectors (see the title of sub-figures). We interpret the results in three perspectives:\\
\indent\textbf{Effect of loop data.} When the probe vehicle ratio is fixed to 0.5\%, the performance of all models are significantly improved as the loop number increases from 0 to 2, which is because the loop data provides more information. This improvement is not significant when the probe vehicle ratio is 3\%, as the probe data alone has provided sufficient information.\\
\indent\textbf{Effects of using FDL.} When the loop number is 2 and the probe vehicle ratio is 0.5\%, PI-GAN-FDL achieves significantly lower REs and KLs compared to PI-GAN and PI-Flow, while this advantage becomes less significant when the data is sparse. This is because the ML surrogate requires more data to train. 
Also, PI-GAN-FDL achieves lower KLs than PI-GAN in general, indicating that PI-GAN-FDL can better capture uncertainty.  \\
\indent\textbf{Comparison between the ARZ-based model and the LWR-based model. } The ARZ-based model outperform the LWR-based model in general, which shows that the second-order physics is more suitable for the real-world scenario. \\
\indent\textbf{Comparison between PIDL-UQ models.} As the data amounts increase, the performance improves and the performance difference across models becomes small.  Among all PIDL based models, TrafficFlowGAN generally achieves the least error in terms of both RE and KL, because it combines the advantages of both PhysGAN and PhysFlow. In Fig.~\ref{fig:results:ngsim_sto}(e) we summarize the $RE$ (x-axis), i.e. the relative difference between the predicted and ground-truth mean, and $KL$ (y-axis), i.e. the statistical 
difference between the predicted and ground-truth distribution, of all models with different training datasets that are shown in Fig.~\ref{fig:results:ngsim_sto}(a-d). Each point represents a combination of metrics by applying one model type to one training dataset. We interpret these points by assigning them into 4 regions:
\begin{itemize}
    \item Region A (optimal RE and KL): Most points in this region belong to the  TrafficFlowGAN model type (in stars), which shows that the combination of PI-GAN and PI-Flow help achieve the best performance in terms of both RE and KL.
    \item Region B (low RE and high KL): Most points in this region belong to GAN (in upsided triangle) and PI-GAN (in dot), which is a sign that the GAN-based models are prone to mode-collapse. 
    \item Region C (balanced RE and KL): Most points in this region belong to the PI-Flow model type, indicating that explicit estimation of data likelihood helps balance RE and KL. 
    \item Region D (high RE and low KL): Most points in this region belong to the EKF (in triangle) and PI-GAN-FDL (square), showing that these two types of models can better capture the uncertainty than the mean.
\end{itemize}

\begin{figure}[h]
\centering
  \includegraphics[scale=0.3]{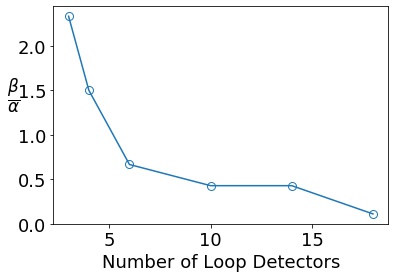}
  \caption{Ratios of the contributions made by the physics-based component and the data-driven component to the optimal training of TraifficFlowGAN. $\beta$ and $\alpha$ are hyperparameters in Eq.~\ref{equ-3-B-2}, which control the contribution of physics-based and data-driven components, respectively. }
  \label{fig:beta-alpha-stoc}
\end{figure}

\textbf{Transition from pure physics-driven to data-driven TSE models}: Fig.~\ref{fig:beta-alpha-stoc} shows the optimal $\beta/\alpha$ ratio of the TrafficFlowGAN under different numbers of loop detectors. The optimal $\beta/\alpha$ has a similar decreasing trend as in the the deterministic TSE problem shown in Fig.~\ref{fig:beta-alpha-deter}.

\section{Conclusion and Future Work}\label{sec:conclusion}

\subsection{Conclusions}

This paper lays a methodological paradigm of PIDL for the TSE problem, including traffic state inference and uncertainty quantification. 
We present the concept of HCG that integrates both a PUNN and a PICG. 
In the TSE problem in particular, we present various architecture design. 
For the traffic state inference problem, the PUNN and PICG can be linked in sequence or in parallel, depending on whether traffic velocity is computed from traffic density using FDs, or whether both velocity and density are computed from PUNN(s) simultaneously. 
For the UQ problem, GAN and non-GAN based adversarial generative models are presented to characterize the impact of measurement uncertainty on traffic states.    
The architecture variants, including PI-GAN, PID-GAN, Mean-GAN, PI-GAN-FDL, PI-Flow, and TrafficFlowGAN are introduced. 
This is the first study that compare PIDL model variants using the same real-world dataset, which provides a benchmark platform for model comparison. 
By comparing various PIDL models, we demonstrate that this paradigm is advantageous over pure physics or data-driven approaches in the ``small data" regime, and also allows a smooth transition from pure physics-driven to data-driven models via tuning the hyperparameters in the loss.
Tab.~\ref{tab:joint_seq_par} summarizes the existing types of hybrid architecture used in PIDL-TSE.
		
		
		
		
		

\begin{table}\centering
\begin{threeparttable}
	\centering
	\caption{Configuration of physics in the physics-based component. \label{tab:joint_seq_par}}
	\begin{tabular}{ |m{1 cm}<{\centering} ||p{1.8 cm}<{\centering} |p{1.8 cm}<{\centering} |}
		\hline
	    \multicolumn{1}{|@{}l||}{\backslashbox[0pt][l]{\makecell[l]{$\hspace{1em}\text{PUNN-}$\\$\hspace{1em}\text{PICG topology}$}}{\makecell[r]{$\hspace{-6em}\text{PUNN}$\\ $\hspace{-6em}\text{hierarchy}$} }} & Shared & Separate  \\ \hline\hline
		Sequential & \cite{shi2021physics_arxiv, shi2021aaai,Liu-2021-IFAC-PapersOnLine, Huang-Jiheng-2020,daw2021pid,yang2019adversarial,yang2019highly,mo2022quantifying,mo2022trafficflowgan} & \cite{shi2021physics, Barreau-2021-CDC, Barreau-2021-CLDC,guo2022normalizing,yang2019adversarial,zhang2019quantifying,yang2020physics}\\
		\hline
		Parallel & \cite{mo2021physics,mo2022uncertainty} & -\\
		\hline
		\end{tabular}
    \end{threeparttable}
    \begin{tablenotes}
      \footnotesize  
      \item Note. PUNN-PICG topology refers to the layout between PUNN and PICG, which can be either ``sequential" (i.e., The output of PUNN is the input into PICG, so PICG follows PUNN.) or ``parallel" (i.e., PUNN and PICG output predictions side-by-side to compute a loss.).
      PUNN hierarchy refers to the layout of NNs 
      that are used to predict $\rho$ and $u$, respectively. ``Shared" means that only one NN is used to output both $\rho,u$, while ``separate" means that two NNs are used to output $\rho,u$, respectively.
    \end{tablenotes}
    \vspace{-2.0em}
\end{table}

\subsection{Outlook}

Looking forward, we will pinpoint several promising research directions that hope to guide researchers to exploit this under-tapped arena. 

\subsubsection{Physics representation}

What physics model is selected depends on data fidelity, model fidelity, and available computational resources.


While there are a significant amount of models to describe traffic dynamics on micro- \cite{treiber2000congested,kesting2008adaptive}, meso- \cite{zhou2014dtalite,di2016network}, and macro-scale \cite{lighthill1955kinematic,Richards-1956,Aw-2002,zhang-2002},  
the future is in the discovery of a multiscale traffic model (i.e., a mapping from multiscale measurements to traffic states on different scales) that collectively infers traffic states with various measurements.  
Analytical multiscale models have been thoroughly studied in fields like  biological \cite{alber2019integrating} and materials science \cite{chinesta2018virtual},
but remain under-exploited in transportation modeling. 
The drivers of developing a multiscale traffic model are two-fold: 
(1) Sensor data are at different scales. 
The collected traffic data include high-resolution individual trajectories and low-resolution aggregate information, both at various spatiotemporal granularity. 
Traffic measurements on different scales essentially measure the same system and phenomenon. 
Accordingly, a multiscale model capable of integrating measurements of various scales could characterize traffic dynamics more accurately with a smaller dataset. 
(2) Traffic optimization and management strategies need to rely on diverse models and data of different scales. 
Individual traces are normally modeled using ODEs while aggregate traffic patterns are modeled with PDEs. 
An integrated ODE-PDE physics model can potentially accommodate these measurements at both micro- and macro-scale \cite{delle2019traffic}. 
Multiscale traffic models could also help reduce model complexity and speed up simulation for real-time applications. 

A multiscale traffic model, however, presents computational challenges due to the curse of dimensionality (i.e., high-dimensional input-output pairs). 
It is thus important to utilize reduced modeling techniques and multi-fidelty modeling \cite{peherstorfer2018survey,penwarden2022multifidelity}. 

An emerging direction to improve physics representation is to consider proxy models. 
For example, symbolic regression has been demonstrated to learn relatively simple physical forms to describe complex physical systems \cite{schmidt2009distilling,cranmer2020discovering}.
Traffic engineers have spent decades discovering various physical laws to describe human behavior models, from car-following, lane-change, to other driving scenarios and tasks. 
Is it possible to develop a systematic method to select mathematical models?
For example, selection of a model can be reformulated as finding an optimal path over the PICG from inputs to outputs \cite{wang2019cooperative}. Accordingly, model selection is converted to an optimization problem.
Machine learning methods, such as neural architecture search 
\cite{elsken2019neural} and automatic modularization of network architecture \cite{chen2020modular}, could enable the automatic architecture design of PICG and PUNN.

\subsubsection{Learning discontinuity in patterns}

Traffic patterns, especially congested traffic, are highly driven by underlying physics, 
thus, how to learn patterns around shockwave, which corresponds to the discontinuity of which gradients do not exist, remains an active research area in PIDL.
There is a small amount of literature that discussed such a limitation using PIDL for nonlinear PDEs with solutions containing shocks and waves. 
One solution is to add viscosity terms \cite{fuks2020limitations,shi2021aaai,shi2021physics} to smoothen the discontinuity.

Because the state-of-the-art primarily focuses on the ``soft" method, which imposes the physics constraints as part of the loss function, the fulfillment of physics constraints cannot be guaranteed, leading to poor prediction in shocks and waves.
Thus, ``hard" methods that enforce physics can be a radical remedy. 
One option is to convert the TSE problem using varational formulation \cite{daganzo2005variational}, and define an energy-based loss function \cite{samaniego2020energy} that naturally adopts the objective function in the variational formulation.
 
\subsubsection{Transfer and meta- learning}
For engineers, an important question is, if we have trained a PUNN using data collected from city $A$, can we directly generalize the prediction out of the NN using data from city $B$? 
Traffic datasets from two cities could differ drastically in the underlying traffic dynamics (arising from heterogeneity in driver behavior such as that in San Francisco and Mumbai), traffic environments, 
road conditions,
initial and boundary conditions, 
and traffic demands. 
To address this challenge, we have to modify the input of the PUNN without using $(x,t)$ but other attributes that vary across datasets, including but not limited to road type, geometry, lane width, lane number, speed limit, travel demands, traffic composition and so on.

Another direction to meet the transfer needs is to ask, how can we train a family of related physics based PDEs (such as LWR and ARZ) and generalize the tuned hyberparameters to other physics members? 
Meta-learning the parameters involved in the PIDL pipeline could be a potential solution \cite{psaros2022meta}.

\subsubsection{IoT data for urban traffic management}

How can we fully exploit the advantage of multimodal, multi-fidelity IoT data, including but not limited to individual trajectories, camera images, radar heatmap, and Lidar cloud points? 
The existing practice is to preprocess these multimodal, multi-fidelity measurements using computer vision
and extract aggregate traffic information in terms of traffic velocity and/or density within discretized spatial cells and time intervals. 
Rather than converting these data formats to conventional ones, 
we should think outside the box and potentially, redefine the entire TSE framework. 
It thus demands a shift in paradigm from what constitutes traffic states to one taking individual trajectories and images as inputs. 
In other words, is it sufficient to simply use aggregate traffic velocity, density, and flux to describe traffic states? 
The aggregate traffic measures were introduced decades ago when inductive loop detectors were deployed to measure cumulative traffic counts.  
Traffic flow has long been regarded analogous to fluids, and accordingly, hydrodynamic theory is applied to model traffic flow dynamics. 
It is natural to adapt physical quantities defined for fluids to traffic flow. 
However, traffic systems are not physical but social systems, in which road users constitute a complex traffic environment, 
while interacting continuously with the built environment. 
Contextual information greatly influences driving behaviors and traffic dynamics. 
Accordingly, the question is, when we describe the state of traffic and aim to optimize traffic management strategies, 
would traffic contextual information perceived by our eyes and brains (represented by DNNs) be helpful as an addition to those widely used quantitative measures, especially when this type of information becomes more widely available, thanks to IoT and smart city technology? 

Furthermore, if TSE models can be enriched with multimodal data, what are new challenges and opportunities to traffic control and optimization models that rely on traffic state inference? 
There are extensive studies on causal discovery and causal inference using observational data when unobserved counfounders are present \cite{spirtes2000causation,pearl2009causality}.
However, little work has been done to leverage explicit causal relations from physical knowledge to improve PIDL. 
For example, counterfactual analysis of physical dynamics concerns identifying the casual effects of various interventions, including traffic control and the sequential decision-making of other agents in the environment \cite{meng2022physics,ruan2022aaai,ruan2022causal}. 
Without performing extra experimentation that is risky and unsafe, how can we design and validate traffic management strategies 
using new information provided by IoT data? 

\subsubsection{TSE on networks}

With the ubiquitous sensors in smart cities, traffic state estimation on large-scale road networks will be more feasible and useful to perform. 
To generalize from single road segments to networks, 
the challenge lies in the spatial representation of graphs, as well as temporal evolution of traffic dynamics.
When PIDL is applied when there is sparse networked sensing information, we need to straighten out in what representation existing physics models could be incorporated, in other words, how we should encode traffic states on links and those at junctions into what deep learning models. 
\cite{ji2022stden} predicts network flows using a spatio-temporal differential equation network (STDEN) that integrates a differential equation network for the evolution of traffic potential energy field into DNNs.

There are a lot more open questions that remain unanswered. 
As this field continues growing, we would like to leave them for readers to ponder:
\begin{enumerate}
    \item How do we leverage various observation data to fully exploit the strengths of PIDL? 
    \item What types of sensors and sensing data would enrich the application domains of PIDL and better leverage its benefits? 
    \item Would there exist a universal architecture of hybrid computational graphs across domains?
    \item What are robust evaluation methods and metrics for PIDL models against baselines? 
\end{enumerate}


\ifCLASSOPTIONcaptionsoff
  \newpage
\fi

\bibliographystyle{IEEEtran}

\bibliography{survey_rongye,survey_Zhaobin, survey_Di}

\end{document}


%

\title{Physics-Informed Deep Learning For Traffic State Estimation: 
\tcb{Challenges, Opportunities, and the Outlook}}

\section{Supplementary Materials}\label{sec:supp}


\subsection{Experimental details for TSE and system identification using loop detectors}
\subsubsection{Experimental configurations}
\label{append:numerical-tse}
The PUNN $f_{\theta}(x,t)$ parameterized by $\theta$ is
designed as a fully-connected feedforward neural network with
8 hidden layers and 20 hidden nodes in each hidden layer. The specific structure is:  layers = [2, 20, 20, 20, 20, 20, 20, 20,
20, 1], meaning that PUNN inputs $(x,t)$ to estimate density $\rho$.
Hyperbolic tangent function (tanh) is used as the activation
function for each hidden neuron in PUNN. In addition, the PICG part in the PIDL architecture is parameterized by physics parameters $\lambda$. \\ \indent
We train the PUNN and identify $\lambda$ through the PIDL architecture using the adaptive moment estimation (Adam) optimizer~\cite{Kingma-2015} for a rough training for about 1,000 iterations. A follow-up fine-grained training is done by the limited-memory Broyden–Fletcher–Goldfarb–Shanno (L-BFGS) optimizer~\cite{Byrd-1995} for stabilizing the convergence, and the process terminates until the loss change of two consecutive steps is no larger than $10^{-16}$. This training process converges to a local optimum $(\theta^*,\lambda^*)$ that minimizes the loss. \\ \indent
For a fixed number of loop detectors, we use grid search for hyperparameter tuning by default. Specifically, since Adam optimizer is scale invariant, we fix the hyperparameter $\alpha$ to 100 and tune the other hyperparameters from [1, 10, 50,100, 150, 200] with some follow-up fine tuning. \\ \indent
The training details are presented in Algorithm~\ref{alg-deterministic}.
\begin{algorithm}[h]
\caption{PIDL training for deterministic TSE.}
\label{alg-deterministic}
\textbf{Initialization}:\\
Initialized PUNN parameters $\theta^0$;
Initialized physics parameters $\lambda^0$;
Adam iterations $Iter$;
Weights of loss functions $\alpha$, $\beta$, and $\gamma$.\\
\textbf{Input}: The observation data ${\cal O}=\{(x^{(i)},t^{(i)},\hat{\rho}^{(i)})\}_{i=1}^{N_o}$;
collocation points $ {\cal C}=\{ ({x^{(j)},t^{(j)}})\}_{j=1}^{N_c}$;  boundary collocation points ${\cal C}_B$, e.g., ${\cal C}_B=\{(0,t^{(i_b)})\}_{i_b=1}^{N_b} \cup \{(1,t^{(i_b)})\}_{i_b=1}^{N_b}$
\begin{algorithmic}[1] 
\label{alg:train-deterministic}
\STATE $k  \leftarrow 0$
\STATE $\tilde{\theta}^0 \leftarrow (\theta^0,\lambda^0)$
\WHILE{$k <Iter$}
\STATE Calculate $Loss$ by Eq.~10 using ($\alpha,\beta,\gamma$) on (${\cal O}$,${\cal C}$,${\cal C}_B$) \\
\STATE $\tilde{\theta}^{k+1} \leftarrow \tilde{\theta}^{k} - \text{Adam}(\tilde{\theta}^{k}, \nabla_{\tilde{\theta}}Loss )$ \\
{\mycommfont{// use Adam for pre-training}}\\
\STATE $k  \leftarrow k+1$
\ENDWHILE
\WHILE{$\tilde{\theta}^k$ not converged}
\STATE Calculate $Loss$ by Eq.~10 using ($\alpha,\beta,\gamma$) on (${\cal O}$,${\cal C}$,${\cal C}_B$) \\
\STATE $\tilde{\theta}^{k+1} \leftarrow \tilde{\theta}^{k} - \text{L-BFGS}(\tilde{\theta}^{k}, \nabla_{\tilde{\theta}}Loss )$ \\
{\mycommfont{// use L-BFGS for fine-grained training}}\\
\STATE $k  \leftarrow k+1$
\ENDWHILE
\RETURN $\tilde{\theta}^k$
\end{algorithmic}
\end{algorithm}
\subsubsection{Additional experimental results using numerical data for three-parameter-based LWR}
The results under different numbers of loop detectors $m$ are shown in Table~\ref{tab:ch5}.
\begin{table}[h!]
    \caption{Prediction and parameter calibration errors of PIDL for the three-parameter-based LWR data }
    \label{tab:ch5}
    \centering
    \begin{threeparttable}
    \begin{tabular}{ccccccc}
    \toprule
    \toprule
    \multicolumn{1}{c}{$m$} &
    \multicolumn{1}{c}{$\text{RE}_{\rho}$(\%)} &
    \multicolumn{1}{c}{$\delta^*$(\%)} &
    \multicolumn{1}{c}{$p^*$(\%)} &
    \multicolumn{1}{c}{$\sigma^*$(\%)} &
    \multicolumn{1}{c}{$\rho^*_{max}$(\%)} &
    \multicolumn{1}{c}{$\epsilon^*$(\%)}\\
    \midrule
    3     & 75.50 & 54.15 & 124.52 & $>$1000 & $>$1000 & 99.95\\
    4     & 10.04 & 59.07 & 72.63 & 381.31 & 14.60 & 6.72\\
     5     & 3.186 & 2.75 & 4.03 & 6.97 & 0.29 & 3.00\\
    6     & 1.125 & 0.69 & 2.49 & 2.26 & 0.49 & 7.56\\
    8     & 0.7619 & 1.03 & 2.43 & 3.60 & 0.30 & 7.85\\
    \bottomrule
    \bottomrule
    \end{tabular}
    \begin{tablenotes}
      \footnotesize
      \item $\lambda^*=(\delta^*,p^*,\sigma^*,\rho^*_{max},\epsilon^*)$ are estimated parameters, compared to the true parameters $\delta=5, p=2, \sigma=1, \rho_{max}=1, \epsilon=0.005$.
      \vspace{-0.5em}
    \end{tablenotes}
  \end{threeparttable}
\end{table}

\subsubsection{Additional experimental results using real-world data}

\begin{figure}
   \centering
   \subfloat[][$SE_{\rho}(x,t)$ of the PIDL-LWR-FDL]{ \includegraphics[width=1.0 \columnwidth]{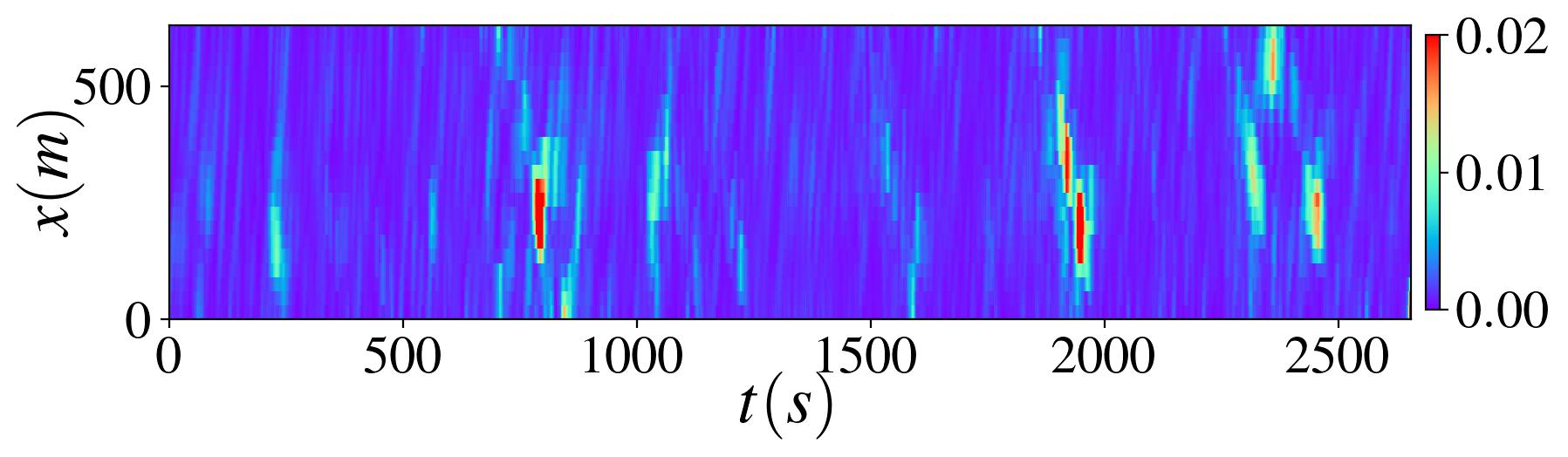}} 
   \\
   \subfloat[][$SE_{\rho}(x,t)$ of the NN]{ \includegraphics[width=1.0 \columnwidth]{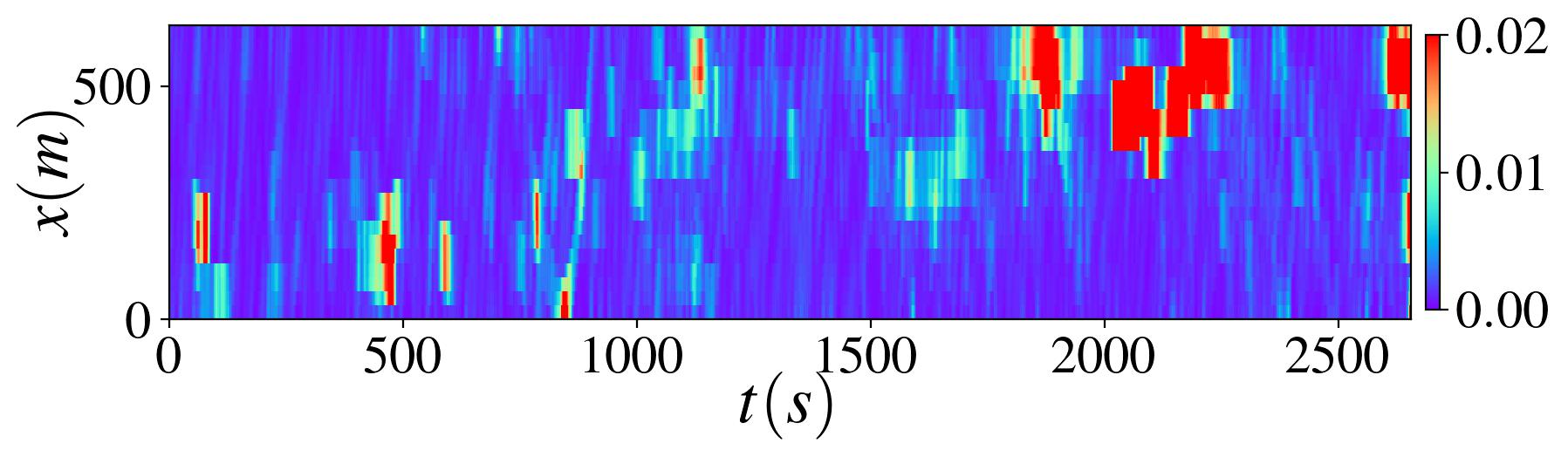}}  
   \caption{Error heatmaps of the NN and PIDL-LWR-FDL models.}
   \label{fig:errormap_deter}
\end{figure}

Figure.~\ref{fig:errormap_deter} shows the error heatmaps for the NN and PIDL-LWR-FDL models when the number of loop detectors is 2 and the prove vehicle ratio is 0.05, where ``SE'' means the squared error.

\subsubsection{Sensitivity analysis on collocation points}
We conduct sensitivity analysis on different numbers of collocation points and the results are presented in Table~\ref{tab:ch5-2}. The detector number is fixed to 5 and hyperparameters remain unchanged. The performances of estimation and parameter discovery improve when more collocation points are used. In this case study, the performance is sensitive to the number of collocations when the collocation rate is smaller than 0.01.
\begin{table}[h!]
    \caption{Sensitivity analysis on collocation rates}
    \label{tab:ch5-2}
    \centering
    \begin{threeparttable}
    \begin{tabular}{p{0.1em}p{0.3em}p{0.3em}p{0.3em}p{0.3em}p{0.3em}p{0.3em}}
    \toprule
    \toprule
    \multicolumn{1}{c}{$C.R.$} &
    \multicolumn{1}{c}{RE$_{\rho}$(\%)} &
    \multicolumn{1}{c}{$\delta^*$(\%)} &
    \multicolumn{1}{c}{$p^*$(\%)} &
    \multicolumn{1}{c}{$\sigma^*$(\%)} &
    \multicolumn{1}{c}{$\rho^*_{max}$(\%)} &
    \multicolumn{1}{c}{$\epsilon^*$(\%)}\\
    \midrule
    0.0001          & 81.2 &         62.66  &   21.15    &   297.77   &    25.91     &         171.37 \\
0.001         &         10.2 &         64.52 &   89.69   &  504.23   &    17.77    &    7.16\\
0.01 &         3.6 &         20.55  &  11.10    &     47.36   &   3.32    &   2.59\\
0.1 &         3.2         & 7.69  &  5.74   &    15.78   &   0.45   &   3.41\\
0.5 &         3.0         & 1.88  &  4.23   &   5.87    &  0.60   &   3.37\\
    \bottomrule
    \bottomrule
    \end{tabular}
    \begin{tablenotes}
      \footnotesize
      \item C.R. stands for the collocation rate, which is the the ratio of the number of collocation points to the number of grid points. The detector number is fixed to 5.
    \end{tablenotes}
  \end{threeparttable}
\end{table}

\subsubsection{Computational effort and model accuracy}
Table.~\ref{tab:compute_time_deter} presents the computational time and prediction error of the deterministic PIDL and baselines with LWR model as the physics when the loop detector number is 2 and the probe vehicle ratio is $3\%$. The first column is the name of the model used in the experiment, and the second column is the type of the model. The third and fourth columns are the computation time for the training and test, respectively. The remaining columns are the performance metrics in terms of prediction error. 

As EKF is not a learning-based method, it can make prediction directly after the parameters of the dynamic model, i.e., discrete LWR or ARZ, are pre-calibrated. Although the deep learning models (PIDL-FDL, NN) require training, their test time is much less than that of the EKF. In terms of accuracy, which is another aspect of evaluating the efficiency of a method, the PIDL and PIDL-FDL models achieve a significant accuracy improvement compared to EKF and NN. The short computational time in the test phase (or in the run time) and high accuracy are both of major consideration in real-world applications.


\begin{table}[h]
    \centering
    \begin{tabular}
    {m{1.6cm}<{\centering} 
    m{1.6cm}<{\centering}
    m{1.cm}<{\centering}
    m{1.cm}<{\centering}
    m{0.4cm}<{\centering}
    m{0.4cm}<{\centering}}
    \toprule
    \toprule
    \multirow{2}{*}{Model}   &
    \multirow{2}{*}{Model Type} &
    \multicolumn{2}{c}{Computation Time} &
    \multicolumn{2}{c}{Prediction Error}\\
    \cmidrule(lr){3-4} \cmidrule(lr){5-6}
     & &\makecell{Training\\(s)} & \makecell{Test\\(s)} &
     \makecell{$\text{RE}_{\rho}$ \\ $(\%)$} &
     \makecell{$\text{RE}_{u}$ \\ $(\%)$}\\
    \midrule
    EKF & Physics-based & $-$ & 1.5 & 39.5 & 35.2 \\
    NN & Data-driven &248 & 0.03 & 42.8 & 39.8 \\
    PIDL  & Hybrid & 647 & 0.04 & 31.5 & 30.5 \\
    PIDL-FDL & Hybrid & 769 & 0.04 & \textbf{21.2} & \textbf{11.6}\\
    \bottomrule
    \bottomrule
    
    \end{tabular}
    \caption{The computation time and prediction error of deterministic PIDL models and baselines}
    \label{tab:compute_time_deter}
    \begin{tablenotes}
      \footnotesize
     \item As EKF is not a learning based method, the ``training time'' is not applicable to EKF.
    \end{tablenotes}
\end{table}

\subsection{Experimental details for UQ-TSE and system identification using loop detectors}

\subsubsection{Experimental configurations}
For the PI-GAN model, the generator structure is: layers = [3, 20, 40, 60, 80, 60, 40, 20, 2], meaning that the generator inputs $(x, t, z)$ to estimate both the traffic density $\rho$ and traffic velocity $u$. The discriminator structure is: layers = [4, 20, 20, 40, 60, 80, 60, 40, 20, 20, 1], meaning the discriminator inputs $(x, t, \rho, u)$ and output a one-dimensional real number indicating whether the input $\rho$ and $u$ are from the real-world data or not. Note that the discriminator has a more complex structure than the generator to stabilize the training. The Rectified Linear Unit (ReLU) is used as the activation function for each hidden layer.

For the PI-Flow model, both the scale neural network and the transition neural network share the same structure, i.e., 2 hidden layers with 64 neurons each layer. The number of transformation is 6, that is, there are 6 scale neural networks and 6 transition neural networks in total. Leaky Relu is used as the activation function for each hidden layer.

For the TrafficFlowGAN, its PUNN is a normalizing flow model that share the same architecture as that in the PI-Flow model introduced above. The discriminator consists of a stack of 3 convolutional layers. The kernel size for each layer is 3, and the number of channels of each layer is 4, 8, and 16, respectively. 

We use Adam optimizer to training the neural networks for 5000 iterations. Different from training PIDL for deterministic TSE, L-BFGS optimizer is not used, because the GAN model requires training the generator and discriminator in turn, for which the L-BFGS optimizer is not suitable. We fix the hyperparameter $\beta=1-\alpha$ and tune $\alpha$ from [0.1, 0.3, 0.4, 0.6, 0.7].

The training details are presented in Algorithm.~\ref{alg:sto}.

\begin{algorithm}[t]
\caption{PIDL-UQ training for stochastic TSE.}
\label{alg:sto}
\textbf{Initialization}:\\ 
Initialized physics parameters ${\lambda}^0$;
Initialized networks parameters $\theta^0$, $\phi^0$;
Training iterations $Iter$;
Batch size $m$;
Learning rate $lr$;
Weights of loss functions $\alpha$, $\beta$, and $\gamma$.\\
\textbf{Input}: The observation data ${\cal O}=\{(x^{(i)},t^{(i)},\hat{\rho}^{(i)})\}_{i=1}^{N_o}$;
collocation points $ {\cal C}=\{ ({x^{(j)},t^{(j)}})\}_{j=1}^{N_c}$;  boundary collocation points ${\cal C}_B$, e.g., ${\cal C}_B=\{(0,t^{(i_b)})\}_{i_b=1}^{N_b} \cup \{(1,t^{(i_b)})\}_{i_b=1}^{N_b}$
\begin{algorithmic}[1] 
\label{alg:train}
\FOR{$k \in \{0,...,Iter\}$}
\STATE Calculate $Loss$ by Eq.~10 and Eq.~11 using ($\alpha,\beta,\gamma$) on (${\cal O}$,${\cal C}$,${\cal C}_B$)  \\
{\mycommfont{// update the generator (PUNN)}}\\
\STATE $\theta^{k+1} \leftarrow \phi^{k} - \text{Adam}(\theta^{k},\nabla_{\theta} Loss)$ \\
\STATE Calculate $Loss_{\phi}$ by Eq.~12\\
{\mycommfont{// update the discriminator (for GAN, PhysGAN, and PhysFlowGAN)}}\\
\STATE $\phi^{k+1} \leftarrow \phi^{k}-\text{Adam}(\phi^{k},\nabla_{\phi} Loss_{\phi})$ \\
\ENDFOR
\end{algorithmic}
\end{algorithm}
\subsubsection{Additional experimental results using numerical data validation for Greenshields-based ARZ}
The results under different numbers of loop detectors $m$ are shown in Table.~\ref{tab:num_physflow_arz}.

\begin{table}
    \caption{Prediction and parameter calibration errors of PI-GAN for the Greenshields-based ARZ data}
    \label{tab:num_physflow_arz}
    \centering
    \begin{threeparttable}
    \begin{tabular}{ccccccc}
    \toprule
    \toprule
    \multicolumn{1}{c}{$m$} & 
    \multicolumn{1}{c}{$\text{RE}_{\rho}$(\%)} & 
    \multicolumn{1}{c}{$\text{RE}_{u}$(\%)} & 
    \multicolumn{1}{c}{$\text{KL}_{\rho}$} & 
    \multicolumn{1}{c}{$\text{KL}_{u}$} & 
    \multicolumn{1}{c}{$u^*_{max}$(\%)} &
    \multicolumn{1}{c}{$\rho^*_{max}$(\%)}\\
    \midrule
    3 & 42.6 & 32.5  & 1.325 & 0.985 & 11.5 & 15.6\\
    4 & 30.2 & 20.9  & 0.965 & 0.835 & 6.5  & 4.5 \\
    6 & 20.6 & 11.8  & 0.753 & 0.638 & 2.9 & 2.2\\
    8 & 18.5 & 6.3   & 0.663 & 0.621 & 2.3 & 1.9\\
    \bottomrule
    \bottomrule
    \end{tabular}
    
    \begin{tablenotes}
      \footnotesize
      \item $m$ stands for the number of loop detectors.  $\lambda^*=(\rho^*_{max},u^*_{max})$ are estimated parameters, compared to the true parameters $\rho_{max}=1.13,u_{max}=1.02$.
    \end{tablenotes}
  \end{threeparttable}
  \vspace{-1.2em}
\end{table}

\subsubsection{Additional experimental results using real-world data}

\begin{figure}
   \centering
   \subfloat[][$SE_{\rho}(x,t)$ of the TrafficFlowGAN]{ \includegraphics[width=1.0 \columnwidth]{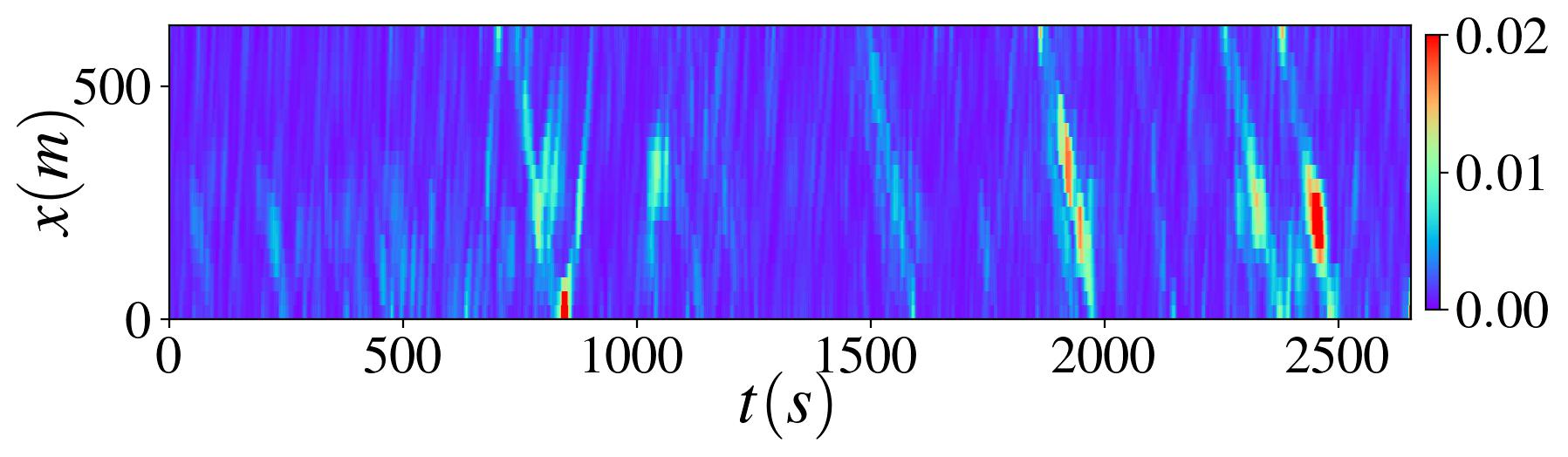}} 
   \\
   \subfloat[][$SE_{\rho}(x,t)$ of the EKF]{ \includegraphics[width=1.0 \columnwidth]{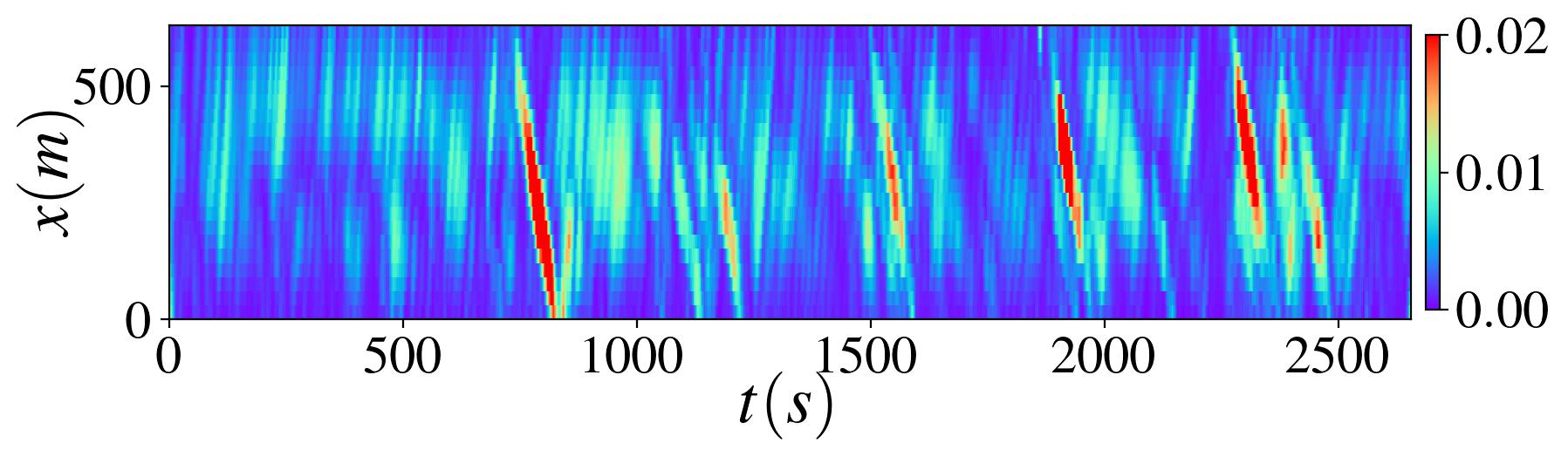}}  
   \\
   \subfloat[][Prediction standard deviation of the traffic density of the TrafficFlowGAN]{ \includegraphics[width=1.0 \columnwidth]{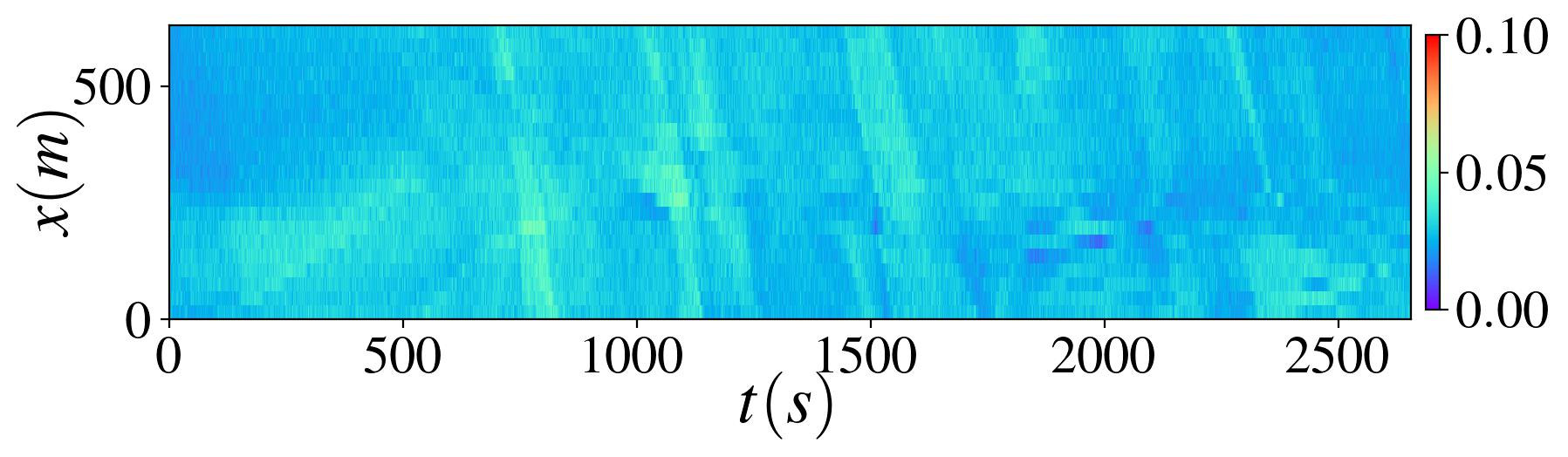}} 
   \\
   \subfloat[][Prediction standard deviation of the traffic density of the EKF]{ \includegraphics[width=1.0 \columnwidth]{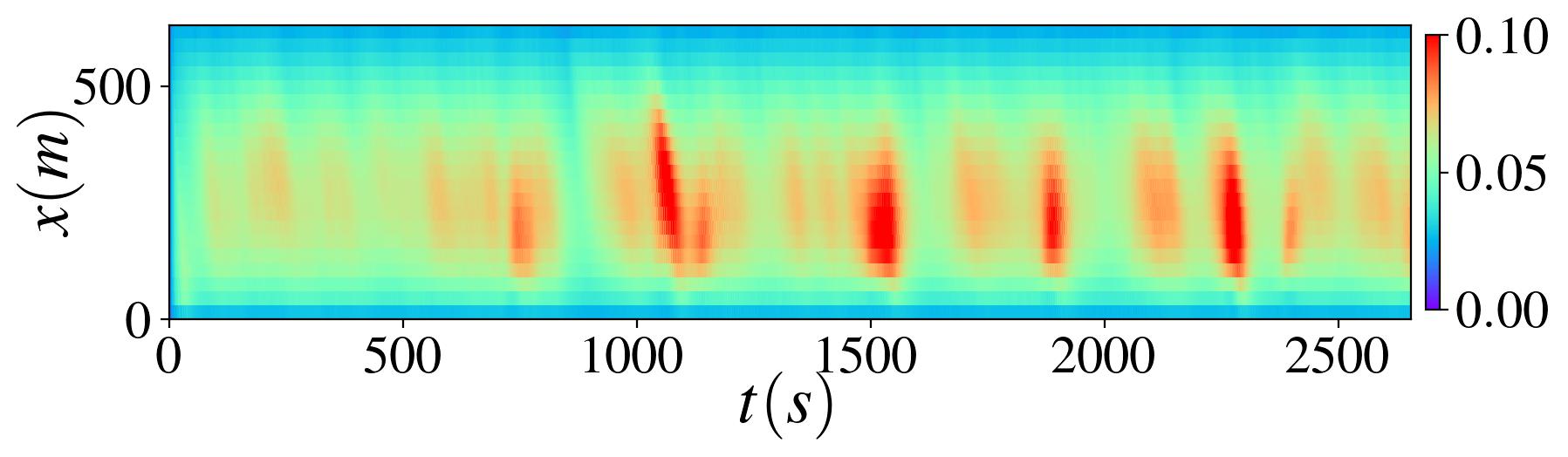}} 
   \caption{Error heatmaps of the EKF and TrafficFlowGAN.}
   \label{fig:errormap_sto}
\end{figure}

Figure.~\ref{fig:errormap_sto} shows the error and prediction standard deviation heatmaps for the EKF and TrafficFlowGAN models when the number of loop detectors is 2 and the prove vehicle ratio is 0.05. 

\subsubsection{Computational time and model accuracy}
Table.~\ref{tab:compute_time_sto} presents the computational time and prediction error of the stochastic PIDL-UQ with LWR model as the physics when the loop detector number is 2 and the probe vehicle ratio is $3\%$. The column specification of which is the same as Table.~\ref{tab:compute_time_deter}, except for that there are more performance metrics for the UQ-TSE problem. The deep learning based models (GAN, PI-GAN, PI-Flow, and TrafficFlowGAN) 
achieve a much less test time compared to the EKF, and there is a significant accuracy improvement for the PIDL based models, especially for the TrafficFlowGAN model. This property is similar as in Table.~\ref{tab:compute_time_deter} when training  and testing the deterministic PIDL models. Among the PIDL based models, PI-Flow has the shortest training time because it uses the normalizing flow model as the generator (PUNN), which can be trained by maximum likelihood estimation and does not require a discriminator. Using both normalizing flow and GAN, TrafficFlowGAN consume the most time for training.

\begin{table}[h]
    \centering
    \begin{tabular}
    {m{1.6cm}<{\centering} 
    m{1.6cm}<{\centering}
    m{1.cm}<{\centering}
    m{1.cm}<{\centering}
    m{0.4cm}<{\centering}
    m{0.4cm}<{\centering}
    m{0.4cm}<{\centering}
    m{0.4cm}<{\centering}
    }
    \toprule
    \toprule
    \multirow{2}{*}{Model}   &
    \multirow{2}{*}{Model Type}   &
    \multicolumn{2}{c}{Computation Time} &
    \multicolumn{4}{c}{Prediction Error}
    \\
    \cmidrule(lr){3-4} \cmidrule(lr){5-8}
    & & \makecell{Training \\ (s)} & \makecell{Test \\ (s)} &
    \makecell{$\text{RE}_{\rho}$ \\ $(\%)$} &
    \makecell{$\text{RE}_{u}$ \\ $(\%)$} &
    \makecell{$\text{KL}_{\rho}$ \\ $(\%)$} &
    \makecell{$\text{KL}_u$ \\ $(\%)$}
    \\
    \midrule
    EKF & Physics-based & $-$ & 1.5 & 39.5 & 35.2 & 3.00 & 3.32 \\
    GAN & Data-driven & 3235 & 0.03 & 30.1 & 39.3 & 5.45 & 5.12\\
    PI-GAN & Hybrid & 3326 & 0.03 & 21.1 & 27.8 & 4.08 & 4.02 \\
    PI-GAN-FDL & Hybrid & 3453 & 0.03 & 15.4 & 21.7 & 2.33 & 2.57 \\
    PI-Flow & Hybrid & 2548 & 0.02 & 22.8 & 27.7 & 3.90 & 4.00 \\
    TrafficFlowGAN & Hybrid & 4323 & 0.02 & \textbf{15.2} & \textbf{20.3} & \textbf{2.27} &\textbf{2.07} \\
    \bottomrule
    \bottomrule
    
    \end{tabular}
    \caption{The computation time and prediction error of stochastic UQ-PIDL models and baselines}
    \label{tab:compute_time_sto}
    \begin{tablenotes}
      \footnotesize
     \item As EKF is not a learning based method, the ``training time'' is not applicable to EKF.
    \end{tablenotes}
\end{table}
\subsection{Performance metrics}
We use four different metrics to quantify the performance of our models. In the formulas below, $s(x,t)$ represents the predicted traffic state (i.e., traffic density $\rho$ or traffic state $u$) at location $x$ and time $t$, and $\hat{s}$ represents the ground truth.\\

\begin{enumerate}
    \item \textbf{Relative Error (RE)} measures the relative difference between the predicted and ground-truth traffic states. For the deterministic TSE, RE is calculated by: 
    \begin{equation*}
        \text{RE}_{s}(x,t)=\frac{\sqrt{\sum_x \sum_t (s(x,t)-\hat{s}(x,t))^2}}{\sqrt{\sum_x \sum_t\hat{s}(x,t)^2}}.
    \end{equation*}
    
    \item \textbf{Squared Error (SE)} measures the squared difference between the predicted and ground-truth traffic states at location $x$ and time $t$. For the deterministic TSE, SE is calculated by:
    \begin{equation*}
        \text{SE}_{s}(x,t)={ (s(x,t)-\hat{s}(x,t))^2}.
    \end{equation*}
    
    \item \textbf{Mean Squared Error (MSE)} calculates the average SE over all locations and time, which is depicted as follows:
    \begin{align*}
        MSE_s(x,t) &=  \sum_x \sum_t SE_s(x,t) / N
    \end{align*}
    where N is the total number of data.
    
    \item \textbf{Kullback–Leibler divergence ($\text{KL}$)} measures the difference between two distributions $P(x)$ and $Q(x)$, which is depicted as follows:
    \begin{equation*}
        \text{KL}(P \| Q)=\sum_{x} P(x) \log \left(\frac{P(x)}{Q(x)}\right).
    \end{equation*}
    
\end{enumerate}

Note that for UQ-TSE, the traffic state $s(x,t)$ follows a distribution, and the aforementioned errors measure the difference between the predicted the the ground-truth mean instead. For example, RE and SE for the UQ-TSE are defined as:
    \begin{equation*}
        \text{RE}_{s}(x,t)=\frac{\sqrt{\sum_x \sum_t (\mathbb{E}[s(x,t)]-\mathbb{E}[\hat{s}(x,t)])^2}}{\sqrt{\sum_x \sum_t\mathbb{E}[\hat{s}(x,t)]^2}}.
    \end{equation*}
    \begin{equation*}
    \text{SE}_{s}(x,t)={ (\mathbb{E}[s(x,t)]-\mathbb{E}[\hat{s}(x,t)])^2}.
    \end{equation*}

\bibliographystyle{IEEEtran}
\bibliography{survey_rongye,survey_Zhaobin, survey_Di}






